\documentclass{article} 
\usepackage{iclr2022_conference,times}

\usepackage{amsmath,amsfonts,bm}

\def\eqref#1{equation~\ref{#1}}

\def\1{\bm{1}}

\DeclareMathAlphabet{\mathsfit}{\encodingdefault}{\sfdefault}{m}{sl}
\SetMathAlphabet{\mathsfit}{bold}{\encodingdefault}{\sfdefault}{bx}{n}

\DeclareMathOperator*{\argmin}{arg\,min}

\usepackage[normalem]{ulem}

\usepackage{hyperref}
\usepackage{url}
\usepackage[pdftex]{graphicx}
\usepackage{multirow}
\usepackage{booktabs}       
\usepackage{caption}

\usepackage{algorithm}
\usepackage{algpseudocode}
\algnewcommand\algorithmicforeach{\textbf{for each}}
\algdef{S}[FOR]{ForEach}[1]{\algorithmicforeach\ #1\ \algorithmicdo}
\usepackage{mathtools}

\usepackage{amssymb}
\usepackage{pifont}
\newcommand{\cmark}{\ding{51}}%
\newcommand{\xmark}{\ding{55}}%

\title{Center Loss Regularization for Continual Learning}

\author{Kaustubh Olpadkar \thanks{ Equal contribution}\\
Stony Brook University, NY, USA \\
\texttt{kolpadkar@cs.stonybrook.edu} \\
\And
Ekta Gavas \footnotemark[1]\\
IIIT Hyderabad, India\\
\texttt{ekta.gavas@research.iiit.ac.in} \\
}

\iclrfinalcopy 

\begin{document}

\maketitle

\begin{abstract}
The ability to learn different tasks sequentially is essential to the development of artificial intelligence. In general, neural networks lack this capability, the major obstacle being catastrophic forgetting. It occurs when the incrementally available information from non-stationary data distributions is continually acquired, disrupting what the model has already learned. Our approach remembers old tasks by projecting the representations of new tasks close to that of old tasks while keeping the decision boundaries unchanged. We employ the center loss as a regularization penalty that enforces new tasks' features to have the same class centers as old tasks and makes the features highly discriminative. This, in turn, leads to the least forgetting of already learned information. This method is easy to implement, requires minimal computational and memory overhead, and allows the neural network to maintain high performance across many sequentially encountered tasks. We also demonstrate that using the center loss in conjunction with the memory replay outperforms other replay-based strategies. Along with standard MNIST variants for continual learning, we apply our method to continual domain adaptation scenarios with the Digits and PACS datasets. We demonstrate that our approach is scalable, effective, and gives competitive performance compared to state-of-the-art continual learning methods. 
\end{abstract}

\section{Introduction}

Humans have the ability to continuously evolve, accumulate and transfer acquired knowledge to learn new skills throughout their lifetime. In contrast, in the classical machine learning paradigm, typically referred to as \textit{isolated learning} \citep{chen2018lifelong}, systems are capable of achieving high performance in learning isolated tasks or narrow domains without using previously learned knowledge. This makes them different from real-world settings where systems are expected to learn consecutive tasks with changing data distributions and unknown task boundaries. In this scenario, the intelligent agent should learn continually without forgetting the already acquired knowledge. Thus, continual learning, or traditionally called lifelong learning \citep{chen2018lifelong,thrun1995lifelong,thrun1996learning, thrun1998lifelong, thrun2012learning}, becomes necessary for artificial general intelligence.

A significant problem in continual learning is catastrophic forgetting in neural networks, also known as catastrophic interference \citep{mccloskey1989catastrophic}. The newly learned information may interfere and disrupt the already learned knowledge, leading to a performance loss on old tasks \citep{ ratcliff1990connectionist}. The extent to which the system should be prone to refine and integrate new knowledge and retain previous information was termed as a \textit{stability-plasticity dilemma} and is well-studied in many previous works. \citep{grossberg1982does, grossberg2013adaptive, mermillod2013stability}. This issue of catastrophic forgetting is known to exist in many different types of neural networks, from standard backpropagation networks to unsupervised networks like self-organizing maps \citep{richardson2008critical, mermillod2013stability}. There have been several attempts to overcome catastrophic forgetting in neural networks, and the various approaches are discussed in the next section.

\subsection{Continual Learning Approaches}
In general, current continual learning methods can be broadly categorized into three different types of strategies based on how they attempt to solve the problem of catastrophic forgetting.

\textbf{Architectural approaches} mitigate catastrophic forgetting by modifying the architectural properties of the networks, e.g., adding more neurons or layers or incorporating weight freezing strategies. One of the earliest strategies in this category was Progressive Neural Networks (PNN) proposed by \citet{rusu2016progressive} that retains a pool of pre-trained models as knowledge and learns lateral connections among them to learn the task at hand. Another simpler strategy, Copy Weights with Re-init (CWR), was proposed \citep{lomonaco2017core50} where consolidated knowledge is maintained by isolating the subsets of weights for each class and learning rest task-specific parameters. Scalability remains an issue in this category of approaches as the network parameters explode as the number of tasks increases.

\textbf{Replay or rehearsal-based approaches} maintain a memory buffer with samples from previous tasks to replay with the examples from the current task to strengthen the old memories. \citep{rolnick2018experience}. \citet{lopez2017gradient} proposed Gradient Episodic Memory (GEM) that favors positive backward transfer and hence mitigating forgetting by using episodic memory of samples from previous tasks. Later, \citet{AGEM} proposed a more efficient and faster version, called Averaged GEM (A-GEM).  Inspired by the suggestion that the hippocampus is better paralleled with a generative model than a replay buffer \citep{ramirez2013creating, stickgold2007sleep}, the replay approach was improved further by replacing the memory buffer with a generative model which could generate unlimited pseudo data from past tasks  \citep{shin2017continual, van2018generative}. Instead of using stored input samples, the replay of latent representations to mitigate forgetting is also explored in many recent works \citep{pellegrini2020latent,van2020brain}.

\textbf{Regularization-based approaches} attenuate catastrophic forgetting by imposing constraints on the update of the network weights \citep{parisi2019continual}. It is generally formulated via additional regularization terms that penalize changes in the weights or predictions of neural network. Learning Without Forgetting (LwF)  \citep{li2017learning} distills knowledge with the network's previous version to enforce the predictions of current and previous tasks to be similar.
Many recent regularization based methods apply penalty on network parameters and estimate the importance of different network parameters. \citet{kirkpatrick2017overcoming} proposed the Elastic Weight Consolidation (EWC), which imposes a quadratic penalty on the difference between the old and new task parameters to slow down the learning on certain weights based on their importance for previous tasks \citep{parisi2019continual}. In Synaptic Intelligence (SI) \citep{zenke2017continual}, individual synapses are allowed to estimate their importance by computing the path integral
of the gradient vector field along the parameter trajectory. Whereas Memory Aware Synapses (MAS) \citep{aljundi2018memory} computes importance based on the sensitivity of predicted output function to each parameter. On the other hand, instead of imposing penalty directly on weights, Less-Forgetful Learning (LFL) \citep{jung2018less} regularizes the \(L_2\) distance between the new and old feature representations to preserve the previously learned input-output mappings by computing auxiliary activations with the old task parameters. Recent regularization works also build upon the traditional
Bayesian online learning framework with variational inference \citep{nguyen2017variational, ahn2019uncertainty, Adel2020Continual}.

\subsection{Motivation}
The architectural approaches suffer from scalability issues as the number of parameters increases with the number of tasks \citep{parisi2019continual}. On the other hand, rehearsal-based strategies generally require large memory buffers to store old task data for high performance. Moreover, in real-world scenarios, it is not always possible to have access to old task data. The generative replay-based methods attempt to solve this issue but are often difficult to train and computationally expensive. On the contrary, the regularization-based strategies assume that all the information essential about the old task is contained in the network weights \citep{kirkpatrick2017overcoming}. The over-parameterization in neural networks makes it possible for the solution to a new task to be found close to the solution for the old task \citep{hecht1992theory, sussmann1992uniqueness, kirkpatrick2017overcoming}. Thus, the regularization strategies are generally memory efficient and computationally less expensive than the other two approaches. Our approach belongs to this category as it focuses on alleviating the catastrophic forgetting problem by regularizing the network to project the new task representations close to the old task representations while keeping the decision boundaries unchanged. We achieve this using the center loss \citep{wen2016discriminative} as a regularization penalty to minimize forgetting. We show that our approach successfully prevents catastrophic forgetting in a computationally efficient manner without accessing the data from old tasks.

\subsection{Contributions}

The contributions of this paper are as follows: 
\begin{enumerate}
\item  We propose a novel regularization-based continual learning strategy which we refer to as center loss regularization (CLR).
\item  We compare our approach to different continual learning strategies in domain incremental scenarios and show that our approach is scalable, computationally efficient while storing minimal additional parameters.
\item  We show that our approach gives a competitive performance with the state-of-the-art techniques when applied in continual domain adaptation scenarios.
\end{enumerate}



\section{Center Loss Regularization (CLR)}
\label{sec:clr}

Deep neural networks excel at learning the hierarchical internal representations from raw input data by stacking multiple layers, which allows the system to learn complex function mappings from input to output \citep{6338939}. These representations become increasingly invariant to small changes in the input as we go up the layers towards the output layer by preserving the vital information about the input related to the task \citep{Guest626374}. The portion till the last hidden layer is considered the feature extractor, and the last fully connected layer is regarded as a linear classifier as the features extracted from the feature extractor are usually linearly separable due to the softmax activation in the top layer \citep{wen2016discriminative}. The Less-Forgetful Learning (LFL) approach \citep{Jung2016LessforgettingLI} demonstrated that catastrophic forgetting can be prevented if the representations of the new task are projected close to the learned representations of the old task keeping the decision boundaries unchanged. \cite{ramasesh2020anatomy} empirically demonstrated that if the higher layers are stabilized while learning subsequent tasks, the forgetting can be mitigated significantly.

In our approach, we freeze the weights of the last fully connected classification layer to keep the decision boundaries unchanged similar to LFL. However, it is non-trivial to make the learned features for new tasks localize nearby corresponding old task features in the latent space. The LFL solves it by using the \(L_2\) distance between the current model's features and the computed features using the old task model as a regularization penalty to preserve the previously learned input-output mappings. However, this approach is highly memory-intensive and computationally expensive since it requires storing the entire model trained on the old task and does forward pass on it to compute representations for each novel task. \cite{wen2016discriminative} introduced the center loss and demonstrated that the joint supervision of softmax loss and center loss helps to increase the inter-class dispersion and intra-class compactness of deeply learned features. During the training process, the model also learns the centers for each class features around which the deeply learned features are typically clustered \citep{wen2016discriminative}. We exploit these properties of center loss in building our continual learning strategy. More details on the center loss are provided in the Appendix \ref{Discriminative_Feature_Learning_with_Center_Loss}.

We use the center loss as a regularization penalty along with softmax loss. To enforce the model to learn the features for the new task in the proximity of corresponding old task features, we utilize the already learned class feature centers of the old task instead of storing the old model weights like LFL. While learning the new task, we enforce the new task features to be close to the already learned feature centers using the center loss, making the model project the features of all tasks in the same localized region, clustered around corresponding class feature centers. For the new task, we freeze the class centers and reduce the learning rate of feature extractor parameters to prevent significant changes in it while training with the new task. We also provide the findings of our ablation study in the Section \ref{sec:Results} to analyze the effects of letting the centers and decision boundaries change while learning new tasks. As our method needs to store only the feature centers for each class throughout the lifetime, the memory requirement is significantly lower than other approaches where the agent needs to store the model weights or maintain the replay buffer. The extra memory requirements are discussed in detail in Section \ref{sec:extra-mem}

\begin{equation}
\label{eq:joint_loss}
L_t = L_s + \lambda L_c
\end{equation}

\begin{equation}
\label{eq:center_loss_reg}
L_c = \frac{1}{2} \sum_{i=1}^{m} \|f_{L-1}(x_i; \theta^{(n)}) - c_{y_i}^{(o)}\|^2_2
\end{equation}

\begin{equation}
\label{eq:weights}
\hat{\theta}^{(n)} =  \argmin_{\theta^{(n)}} L_t(x, c^{(o)}; \theta^{(n)}) + R(\theta^{(n)})
\end{equation}

The equation \ref{eq:joint_loss} represents the joint loss, a combination of the softmax loss \(L_s\) and the center loss \(L_c\), which is minimized during training. Minimizing \(L_s\) helps solve the current/new task, whereas the term \(L_c\) helps retain already learned knowledge and avoid catastrophic forgetting. A scalar \(\lambda\) is used for balancing the two loss functions. The equation \ref{eq:center_loss_reg} defines the modified center loss \(L_c\), where \(c^{(o)}\) denotes the learned values of feature centers from old task. The centers are kept frozen during the subsequent tasks. This enforces the new task representations to have the same feature centers as the old task, leading the old task and new task representations to stay within proximity in the feature space, reducing the catastrophic forgetting. We obtain the equation \ref{eq:weights} as final objective function where \(R(\cdot)\) denotes a general regularization term, such as weight decay. Finally, we propose the center loss regularization algorithm, as shown in Algorithm \ref{alg:clr_algo}. $N$ denotes the number of training iterations, $D^{(n)}$ denotes the new task data, \(\theta^{(o)}\) and \(\theta^{(n)}\) denote the network parameters for the old and new task, respectively. Note that we use a single-headed model in our experiments as we primarily target domain-incremental scenario of continual learning where task identity need not be inferred at test time \citep{van2019three}. The system learns to adapt to changing input distributions, but the task structure remains the same.

\begin{algorithm}

\caption{Center Loss Regularization (CLR)}

\textbf{Input:} $\theta^{(o)}, c^{(o)}, N, \mathcal D^{(n)}$ \\
\textbf{Output:} $\hat{\theta}^{(n)}$
\begin{algorithmic}[1]

\State $\theta^{(n)} \gets \theta^{(o)}$  \Comment{initialize weights}
\label{alg:clr_algo}
\State Freeze the weights of the softmax classification layer.
\For{$i \gets 1, \, N$} \Comment{training iteration}
    \ForEach {minibatch $\mathcal B \in \mathcal D^{(n)} $}
        \State Backpropagate and update $\theta^{(n)}$ for mini-batch {$\mathcal B$} to minimize loss $L_t + R(\theta^{(n)})$
   \EndFor
\EndFor
\State $\hat{\theta}^{(n)} \gets \theta^{(n)}$
\\
\Return $\hat{\theta}^{(n)}$

\end{algorithmic}
\end{algorithm}

\section{Experiments}
In this section, we first detail the experimental protocols
for evaluating lifelong learning algorithms (Section \ref{sec:Benchmarks}). We have compared our proposed method against different continual learning methods, which are specified in Section \ref{sec:Methods}. Finally, we report our results in Section \ref{sec:Results}.

\subsection{ Experimental protocols}
\label{sec:Benchmarks}

\paragraph{Permuted \& Rotated MNIST} \citep{kirkpatrick2017overcoming} are variants of the original MNIST dataset \citep{lecun1998mnist}. In Permuted MNIST, a fixed permutation of the image pixels is applied to each task's training and test set.
In Rotated MNIST, each task consists of images rotated by a fixed angle between 0 and 180 degrees. We chose 10000 training and 5000 testing samples for both these variants. Each task has a test set with the same rotation transformation as the training set for that task. In these experiments, the model encounters 10 tasks in sequence, each with a unique rotation angle. We evaluate the model on the test sets of the respective datasets after training on each task. We use the fully connected neural network (MLP) with two hidden layers of 100 units, each with ReLU activation for our experiments. We train the network using Adam optimizer \citep{kingma2014adam} on mini-batches of 64 samples for 1 epoch over training set per task with a learning rate of \(3 \cdot 10^{-3}\) for the first task and \(3 \cdot 10^{-4}\) for subsequent tasks. \\
Our method can also be applied in the supervised continual domain adaptation settings where the model needs to adapt to the new domains without degrading the performance on the previously seen domains \citep{jung2018less, Volpi2021ContinualAO}. 

\paragraph{Digit Recognition} We consider four widely used datasets for digit recognition in continual domain adaptation setting:
MNIST \citep{lecun1998mnist}, SVHN \citep{netzer2011reading}, MNIST-M \citep{ganin2015unsupervised} and SYN \citep{ganin2015unsupervised}. To assess the performance of our approach, we train our network on these datasets, one by one in the sequence MNIST → MNIST-M → SYN → SVHN, considering each dataset as one task. This way, the network adapts to harder domains continually. For this protocol, 10000 training and 5000 testing samples are chosen for each dataset, and all images are resized to 28 x 28 pixels. We convert the MNIST dataset to 3-channel images by repeating the original channel 3 times for compatibility with the remaining datasets.
We use the ResNet18 \citep{he2016deep} architecture and train the network on each domain for 20 epochs, with a batch size of 64. We use Adam optimizer \citep{kingma2014adam} with a learning rate of \( 3 \cdot 10^{-4}\), which is reduced to \(3 \cdot 10^{-5}\) after the first domain.

\paragraph{PACS} dataset \citep{li2017deeper} is typically used
to assess the domain generalization scenarios. It consists of four domains, namely Photo (1,670 images), Art Painting (2,048 images), Cartoon (2,344 images), and Sketch (3,929 images). Each domain contains seven categories. We use this dataset to evaluate our approach in a continual domain adaptation setting. We trained the network in sequence Sketches → Cartoons → Paintings → Photos where images become more realistic with the new domain. For each domain, the dataset is split into 70\% training and the remaining 30\% as testing samples. All images are resized to 224 x 224 pixels and standardized with the mean and standard deviation of the ImageNet \citep{deng2009imagenet} dataset. This is due to the fact that this protocol uses ResNet18 \citep{he2016deep} network trained on ImageNet dataset. We train the network on each domain for 5 epochs, with a batch size of 64. We use Adam optimizer \citep{kingma2014adam} with a learning rate of \( 3 \cdot 10^{-4}\), which is reduced to \(3 \cdot 10^{-5}\)
after the first domain.


\subsection{Methods}
\label{sec:Methods}

We compare the performance of the proposed approach with that of the state-of-the-art regularization-based strategies. First, we test the fine-tuning approach in which a single model is trained across all tasks, which is the naive approach. 
Second, we test the LwF \citep{li2017learning} method, which uses only new task data to train the network while preserving the original capabilities. Further, we compare the performance with EWC \citep{kirkpatrick2017overcoming}, SI \citep{zenke2017continual} and MAS \citep{aljundi2018memory} methods which try to estimate synaptic importance of different network parameters and use that information to regularize the weights. We also test the LFL \citep{jung2018less} method, which tries to position the features extracted by the new network, close to the features extracted by the old network. Moreover, we explore frameworks like Uncertainty-regularized Continual Learning (UCL) \citep{ahn2019uncertainty} and Variational Continual Learning (VCL) \citep{nguyen2017variational} based on variational inference. VCL employs a projection operator through KL divergence minimization. UCL solves the drawbacks of VCL by proposing the concept of node-wise uncertainty. Then, we consider two oracle methods: If we assume access to
every domain at every point in time, we can either train on samples from the joint distribution from the beginning \textit{oracle (all)}, or grow the distribution over iterations  \textit{oracle (cumulative)}. With access to samples from all domains, oracles are not generally exposed to catastrophic forgetting; yet, their performance is not necessarily an upper bound \citep{lomonaco2017core50}.

We also compare our approach with several replay-based strategies. We use the basic experience-replay method as a baseline and examine if using CLR as surrogate loss along with experience replay can enhance the performance. Further, we compare its performance with state-of-the-art memory-based methods like average gradient episodic memory (A-GEM) \citep{AGEM} and GDumb \citep{prabhu2020greedy} methods. We also compare CLR with recent continual domain adaptation technique called domain randomization and meta-learning (Meta-DR) \citep{Volpi2021ContinualAO} for continual domain adaptation experiments. For Permuted and Rotated MNIST tasks, we experiment with four different memory sizes of 10, 20, 50, and 100 examples per task. For the Digits dataset, we experiment with 100, 200, 300 replay examples per task. For the PACS dataset, we experiment with 10, 20, 30 replay examples per task.

The metrics used in our experiments to evaluate all considered methods are the average accuracy (ACC) and backward transfer (BWT).  ACC is the average of the accuracy of the model across all encountered tasks, and BWT represents the amount of forgetting at the end of training on all tasks \citep{lopez2017gradient}. The formulae for the metrics are presented in detail in Appendix \ref{appendix:Appendix B}. Each experiment in the paper is carried out for 5 trials, and the final values are reported as mean and standard deviation of results.

\subsection{Results}
\label{sec:Results}

\begin{figure}
    \centering
    \begin{minipage}{0.48\textwidth}
        \includegraphics[width=1.0\textwidth]{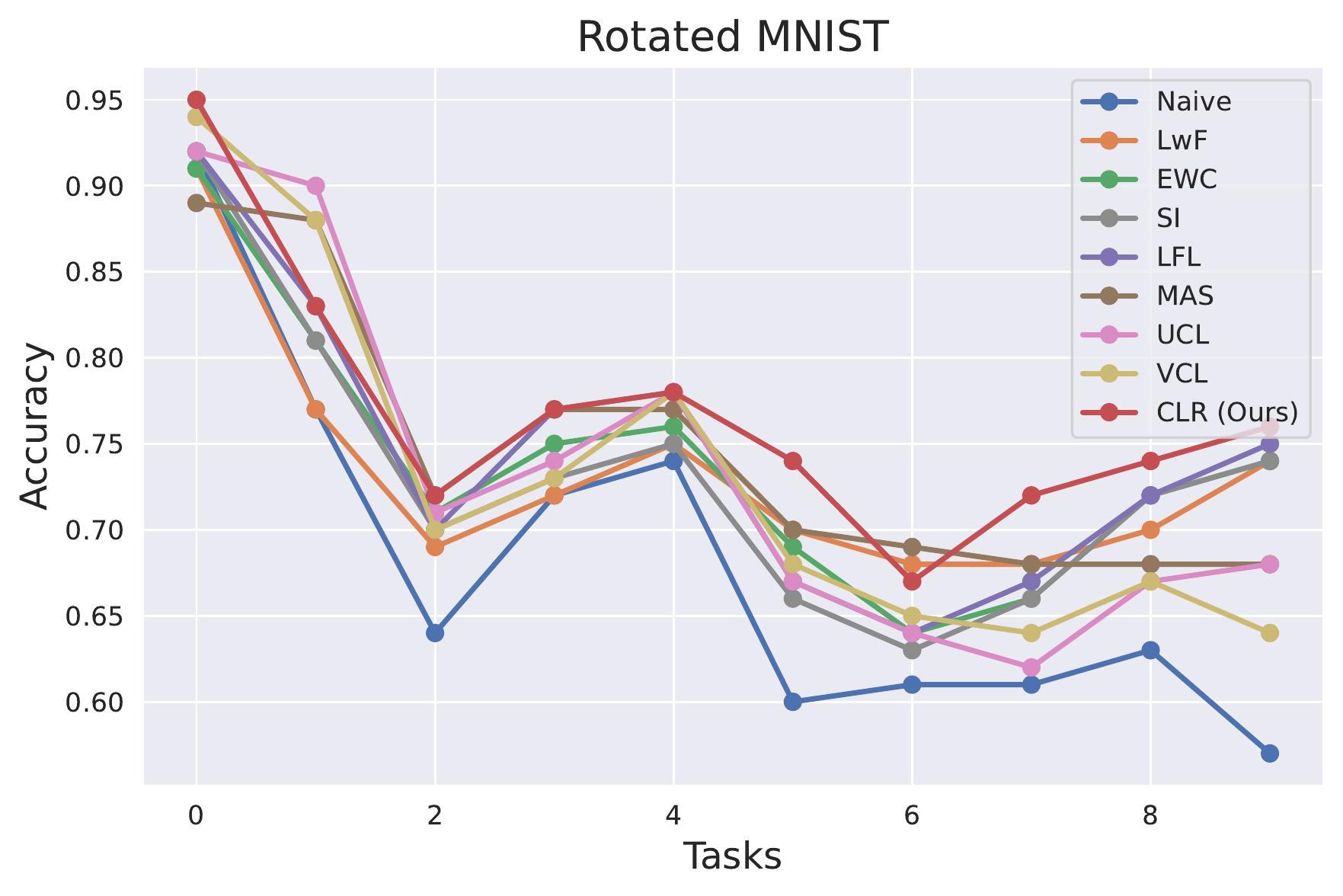}
    \end{minipage}
    \vspace{3mm}
    \centering
    \begin{minipage}{0.48\textwidth}
        \centering
        \includegraphics[width=1.0\textwidth]{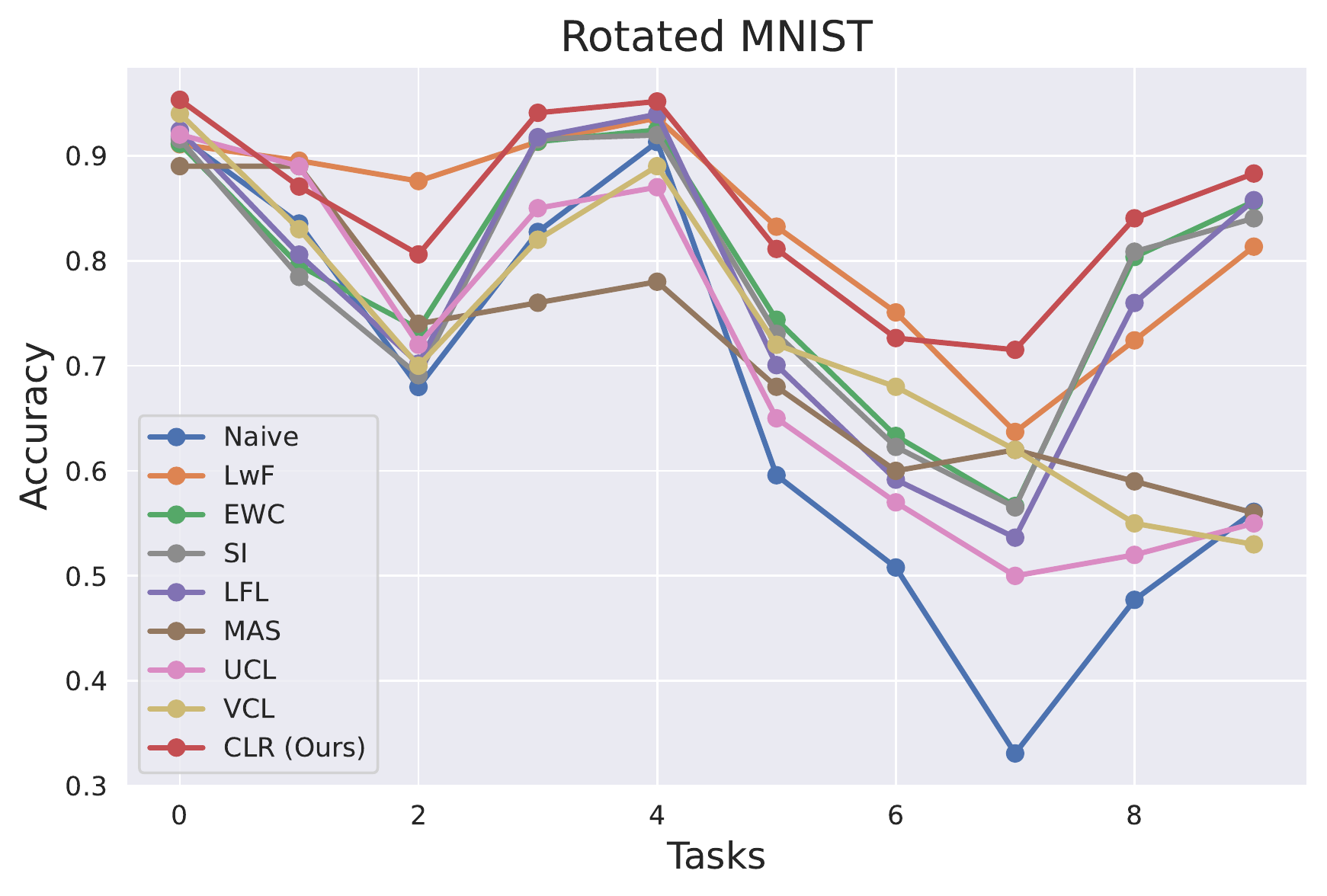}
    \end{minipage}
    
    \centering
    \begin{minipage}{0.48\textwidth}
        \includegraphics[width=1.0\textwidth]{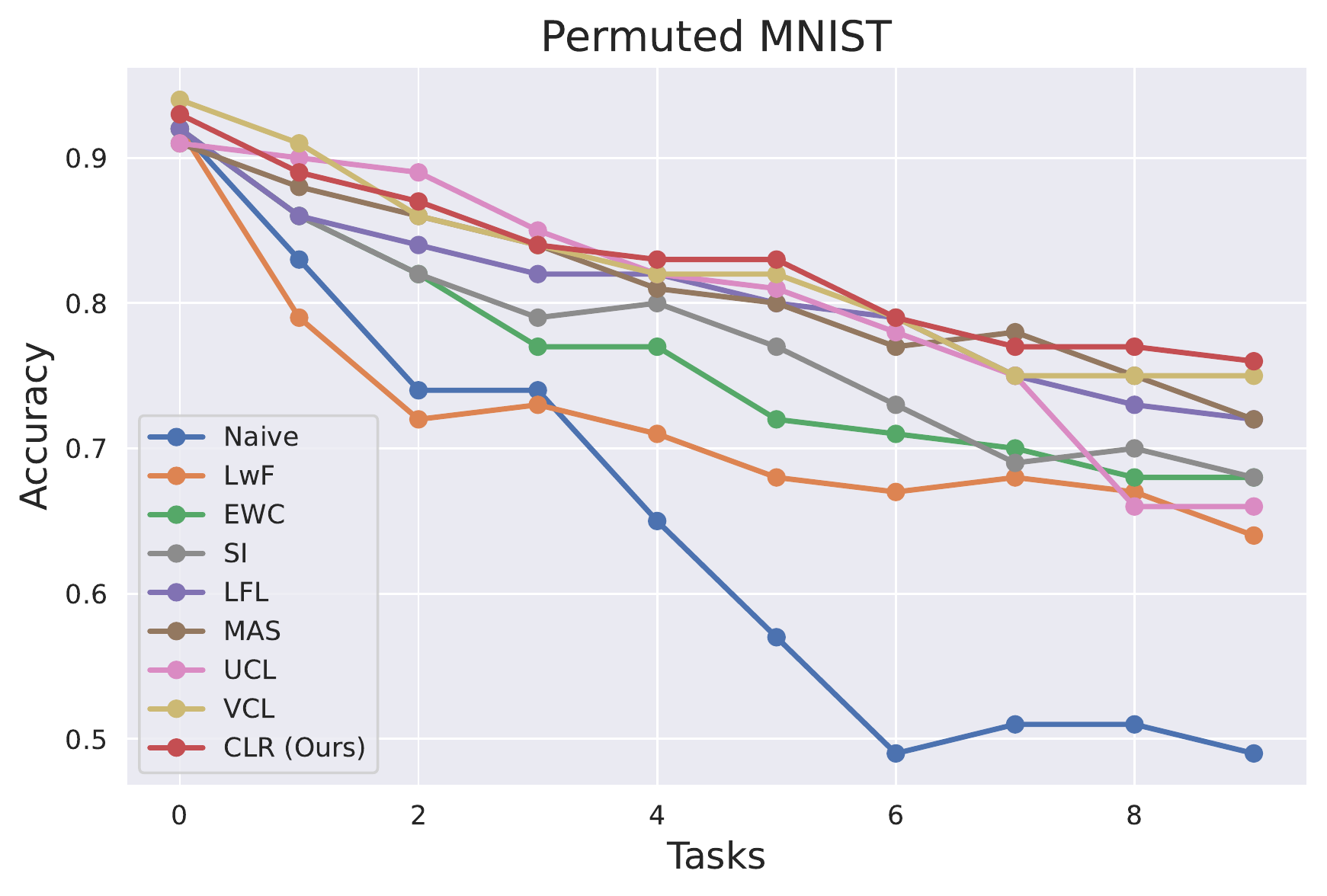}
    \end{minipage}
    \vspace{3mm}
    \centering
    \begin{minipage}{0.48\textwidth}
        \centering
        \includegraphics[width=1.0\textwidth]{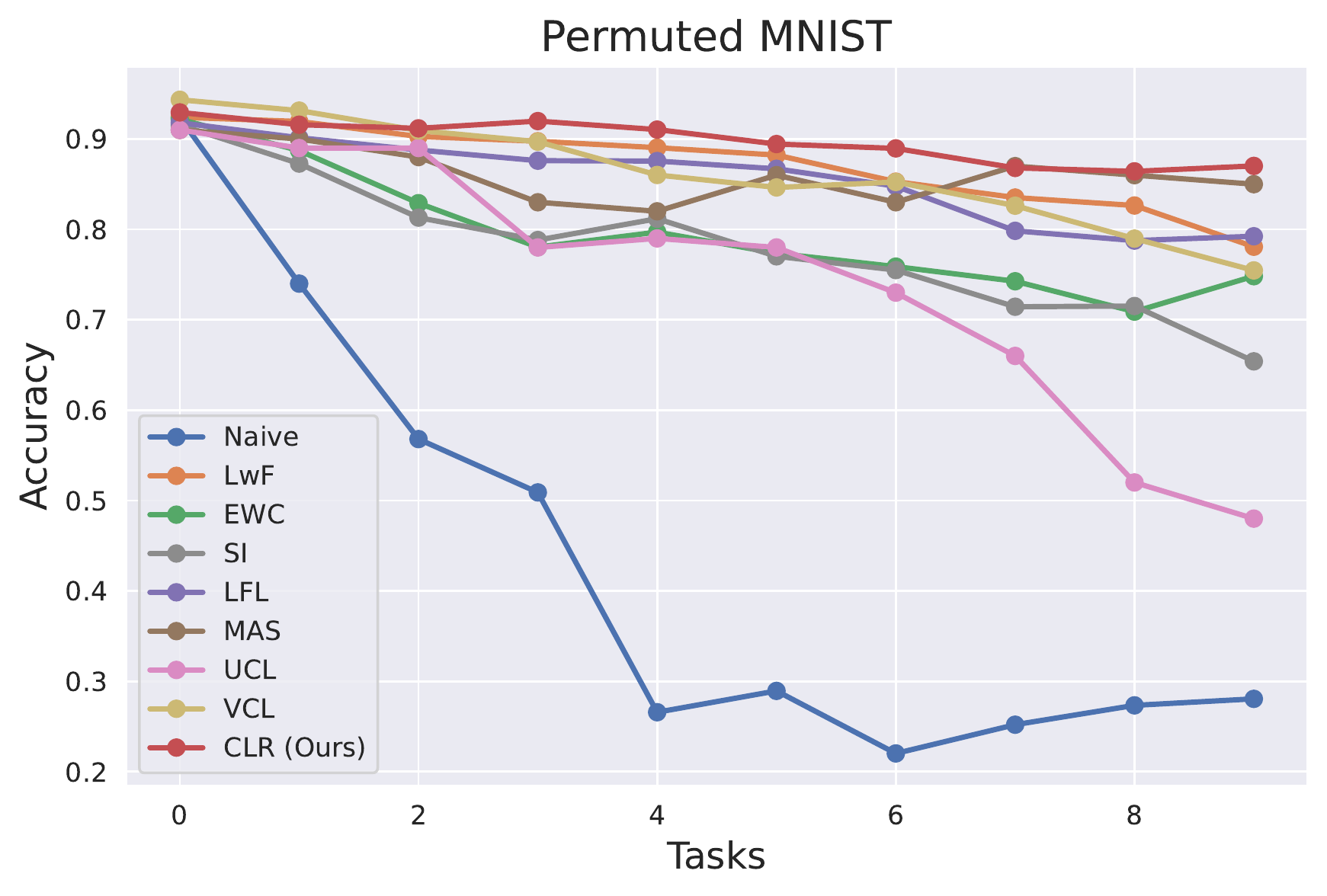}
    \end{minipage}
    
\caption{Left: Average accuracy (ACC) on tasks encountered so far for each method. Right: Progression of the test accuracy for the first task for each method, as more tasks are learned.}
\label{fig:results-mnist}
\end{figure}

Table \ref{table:evaluation-results} compares the performances of different regularization-based approaches on the Rotated and Permuted MNIST datasets. From this table, we can observe that our approach CLR outperforms all other methods. VCL shows competitive performance for Permuted MNIST, but the amount of forgetting (BWT) is worse than CLR. Whereas, LwF shows competitive performance in terms of forgetting but less adaptability to new information. Figure \ref{fig:results-mnist} presents the evolution of the average accuracy (ACC) and the first-task accuracy throughout all the tasks for the MNIST variants. This shows that the center loss regularization helps to mitigate the problem of catastrophic forgetting on earlier tasks while maintaining high performance on all the tasks.

In Table \ref{table:replay-results}, we report the results of replaying samples from episodic memory along with our proposed approach for both MNIST variants. We can observe that using center loss regularization significantly improves the performance over the plain experience replay strategy. Such improvement is consistent as we increase the memory size. It also outperforms the state-of-the-art memory-based methods like A-GEM and GDumb. This shows that our approach can also be used as surrogate loss along with replay-based strategies to enhance performance and reduce forgetting.

We provide the ablation study of our approach in Table \ref{table:abalation-study}. We demonstrate how the performance changes if we do not freeze the centers and classifier weights after the training on the first task is completed. The first row in the table represents the approach proposed in Section \ref{sec:clr}. We try out multiple values of hyperparameter $\lambda$ for each experiment and put the best performing results for each row. These results show that the best performance is generally achieved when we freeze the centers and the decision boundaries after the first task. Moreover, we also demonstrate how the hyperparameter $\lambda$ affects the overall performance of CLR in Appendix Figure \ref{fig:lambda-plots}.

We report in Table \ref{table:domain-adaptation-results-digits} and Table \ref{table:domain-adaptation-results-pacs} the performance of our proposed method compared with different methods in continual domain adaptation setting on Digits and PACS datasets respectively. We note that our approach outperforms the naive, EWC, and SI strategies with a significant margin for both benchmarks. Our method demonstrates competitive performance compared to LwF and LFL strategies. Having access to the samples from older tasks for replay can help reduce catastrophic forgetting significantly compared to regularization methods which do not have access to the data of old domains. Thus, we also examine if using CLR along with Experience Replay (ER) can help boost the performance. For both the benchmarks, we observe that using CLR with ER can significantly improve the overall performance, and it is consistent with the increase in the memory size. The memory size column denotes the number of replay samples per task. These results suggest that the center loss regularization helps the model successfully adapt to new domains without considerable performance degradation on old domains.

\begin{table}[!ht]
\tiny
  \caption{Average test accuracy (in \%) (mean and std) after training on all tasks for each method}
  \label{table:evaluation-results}
\centering
  \begin{tabular}{cccccc}
    \toprule
    \multirow{2}{*}{Method} 
    & \multicolumn{2}{c}{Rotated MNIST} & & \multicolumn{2}{c}{Permuted MNIST} \\
    \cmidrule(r){2-3} \cmidrule(r){5-6}
    
    & ACC & BWT & & ACC & BWT \\
    \midrule
    \midrule
    Naive  & 57.4 ± 1.2 & -30.2 ± 1.5 &&  49.0 ± 1.9 & -47.3 ± 2.2\\
    \midrule
    LwF   & 74.4 ± 0.9 & -12.1 ± 1.3 &&  63.7 ± 0.8 & -12.7 ± 1.1\\
    \midrule
    EWC   & 73.6 ± 0.7 & -16.5 ± 0.8 &&  68.2 ± 0.4 & -19.5 ± 0.3\\
    \midrule
    SI  & 74.4 ± 0.3 & -15.6 ± 0.4 & &  67.5 ± 1.0 & -20.0 ± 1.2\\
    \midrule
    LFL  & 75.5 ± 0.5 & -16.2 ± 1.4 & &  71.9 ± 0.8 & -14.4 ± 1.4\\
    \midrule
    MAS   &  67.5 ± 1.3 & -12.1 ± 1.5 & & 71.8 ± 1.2 & -13.9 ± 1.6\\
    \midrule
    UCL   &  68.2 ± 1.6 & -23.6 ± 1.2 & & 66.5 ± 1.4 & -26.6 ± 1.5\\
    \midrule
    VCL   &  63.6 ± 1.9 & -28.9 ± 1.3 & & 74.6 ± 0.7 & -14.8 ± 1.7\\
    \midrule
    \midrule
    CLR (Ours)  & \textbf{76.4 ± 1.2} & \textbf{-11.9 ± 0.6} &&  \textbf{75.6 ± 1.3} & \textbf{-12.2 ± 0.9}\\
    \bottomrule
  \end{tabular}
\end{table}

\begin{table}[t]
\captionsetup{font=small}
\tiny
\caption{Replay Experiments Study: Average test accuracy (mean \% and std) with varying memory size (Mem size) i.e number of replay examples stored for each task}
  \label{table:replay-results}
\centering
\begin{tabular}{ccccccccccc}
\toprule
\bf Method &
\multicolumn{4}{c}{Rotated MNIST} & & \multicolumn{4}{c}{Permuted MNIST} \\
\midrule
Mem size
& 10 & 20 & 50 & 100 & & 10 & 20 & 50 & 100 \\
\midrule
\midrule
ER & 78.6 ± 1.4 & 81.0 ± 0.5 & 83.3 ± 0.4 & 85.1 ± 0.4 & & 68.9 ± 1.4 & 71.7 ± 1.7 & 80.6 ± 0.4 & 83.5 ± 0.6 \\
\midrule
A-GEM & 77.6 ± 0.4 & 78.5 ± 0.5 & 79.3 ± 0.4 & 79.7 ± 0.6 & & 76.8 ± 0.4 & 77.7 ± 0.5 & 84.3 ± 0.7 & 83.9 ± 0.4 \\
\midrule
GDumb & 78.5 ± 0.3 & 81.1 ± 0.8 & 84.5 ± 0.4 & 85.2 ± 0.5 & & 76.7 ± 0.8 & 78.5 ± 0.9 & 84.3 ± 1.0 & 85.3 ± 0.7 \\
\midrule
\midrule
ER + CLR & \bf 81.0 ± 0.4 & \bf 83.3 ± 0.2 & \bf 85.0 ± 0.2 & \bf 86.6 ± 0.1 & & \bf 78.9 ± 0.5 & \bf 80.0 ± 0.9 & \bf 83.7 ± 0.2 & \bf 86.0 ± 0.4 \\
\bottomrule
\end{tabular}
\end{table}

\begin{table}[!ht]
\captionsetup{font=small}
\tiny
\caption{Ablation Study: Average test accuracy (mean \% and std)}
  \label{table:abalation-study}
\centering
\begin{tabular}{cccccc}
\toprule
\multicolumn{2}{c}{Frozen} & \multicolumn{2}{c}{Rotated MNIST} & \multicolumn{2}{c}{Permuted MNIST} \\

\cmidrule(r){1-2} \cmidrule(r){3-4} \cmidrule(r){5-6}

Centers & Classifier & ACC & BWT & ACC & BWT  \\
\midrule
\midrule
\cmark & \cmark & \textbf{76.4 ± 1.2} & \textbf{-11.9 ± 0.6} & \textbf{75.6 ± 1.3} & \textbf{-12.2 ± 0.9}\\
\midrule
\cmark & \xmark & 76.3 ± 1.2 & -15.0 ± 1.0 & 73.2 ± 0.7 & -15.7 ± 0.8\\
\midrule
\xmark & \cmark & 75.7 ± 0.7 & -15.7 ± 0.9 & 74.3 ± 1.2 & -14.0 ± 1.6\\
\midrule
\xmark & \xmark & 76.4 ± 0.9 & -12.5 ± 0.4 & 73.4 ± 2.4 & -15.4 ± 2.5\\
\bottomrule
\end{tabular}
\end{table}

\begin{table}[!ht]
\captionsetup{font=small}
\tiny
  \caption{Continual Domain Adaptation (Digits): Test accuracy on individual datasets, average test accuracy (ACC) and BWT  (mean \% and std) after training on all domains with Digits protocol}
  \label{table:domain-adaptation-results-digits}
\centering
  \begin{tabular}{cccccccc}
    \toprule
    Method & M.Size & \bf MNIST (1) & \bf MNIST-M (2) & \bf SYN (3) & \bf SVHN (4) & \bf ACC & \bf BWT\\
    \midrule
    \midrule
    Naive & - & 83.4 ± 6.4 & 65.8 ± 3.4  & 71.3 ± 0.4  & 92.6 ± 0.1 & 78.2 ± 0.4 & -12.8 ± 0.7\\
    \midrule
    LwF   & -  & 96.1 ± 1.2  & 72.7 ± 0.9  & 76.6 ± 0.7  & 91.1 ± 1.0 & 84.1 ± 1.3 & -3.6 ± 1.7\\
    \midrule
    EWC   & -  & 93.7 ± 0.8  & 67.9 ± 0.6  & 75.1 ± 0.3  & 92.5 ± 0.2 & 82.3 ± 0.9 & -7.8 ± 0.8\\
    \midrule
    SI  & -  & 91.8 ± 0.5  & 69.2 ± 1.2  & 72.5 ± 0.5  & 91.9 ± 1.1 & 81.3 ± 0.6 & -8.9 ± 0.4\\
    \midrule
    LFL  & -  & 95.5 ± 0.9  & 67.3 ± 0.4  & 77.7 ± 1.3  & 93.2 ± 0.9 & 83.4 ± 0.4 & -7.5 ± 0.3\\
    \midrule
    Meta-DR & -  & 85.7 ± 1.8 & 75.4 ± 0.7 & 82.0 ± 1.9 & 98.5 ± 0.3 & 85.4 ± 1.1 & -8.4 ± 1.0 \\
    \midrule
    CLR  & -  & 95.8 ± 0.6  & 71.9 ± 0.5  &  77.2 ± 0.2 &  93.5 ± 0.2 & 84.6 ± 0.9 & -5.7 ± 0.6\\
    \midrule
    \midrule
     & 100   & 96.2 ± 0.4  & 75.0 ± 1.1  & 79.3 ± 0.3 &  93.3 ± 0.9 & 86.2 ± 0.7  & -4.8 ± 0.7\\
    ER&  200  & 96.7 ± 0.6  & 77.5 ± 1.5  & 80.8 ± 0.7 &  93.7 ± 0.2 & 87.1 ± 0.4  & -3.7 ± 0.8\\
    &  300  & 97.3 ± 1.3  & 78.8 ± 1.2  & 78.6 ± 0.2 &  92.7 ± 0.5 & 86.9 ± 0.5 & -3.6 ± 0.5\\
    
    \midrule
    \midrule
     & 100   & 95.2 ± 0.3  & 77.1 ± 1.3  & 78.6 ± 0.5 &  94.0 ± 0.8 & 85.9 ± 0.5  & -5.0 ± 1.0\\
    A-GEM&  200  & 95.8 ± 1.0  & 77.9 ± 0.8  & 79.1 ± 0.4 &  94.2 ± 0.3 & 86.7 ± 0.6  & -3.9 ± 0.3\\
    &  300  & 96.9 ± 0.8  & 78.7 ± 0.9  & 80.0 ± 1.2 &  93.8 ± 0.6 & 87.3 ± 0.8 & -3.5 ± 0.9\\
        \midrule
    \midrule
     & 100   & 95.4 ± 0.8  & 79.6 ± 0.4  & 80.8 ± 0.3 &  92.7 ± 0.1 & 87.1 ± 0.4  & -3.7 ± 0.3\\
    ER + Meta-DR &  200  & 96.1 ± 0.7  & 80.1 ± 0.9  & 81.2 ± 0.5 &  93.0 ± 0.4 & 87.5 ± 0.6  & -3.4 ± 0.4\\
    &  300  & 97.2 ± 0.4  & 80.3 ± 0.5  & 82.3 ± 0.1 &  93.4 ± 0.3 & 88.3 ± 0.4 & -2.8 ± 0.4\\
    \midrule
    \midrule
    & 100  & 97.1 ± 0.2  & 79.0 ± 0.5  & 81.8 ± 1.4  & 94.3 ± 0.8 & 88.1 ± 0.6  & -2.7 ± 0.3\\
    ER + CLR  &  200 & 97.4 ± 0.5  & 80.8 ± 1.0  & 81.8 ± 0.3  & 94.1 ± 0.9 & 88.7 ± 1.2 & -2.1 ± 0.5\\
     & 300  & 97.9 ± 0.8  & 81.3 ± 1.0  & 82.6 ± 0.3  & 94.4 ± 0.6 & 89.1 ± 0.8 & -1.7 ± 0.9\\
    \midrule
    \midrule
    Oracle (all) & -  &  98.6 ± 0.3  & 90.7 ± 0.4  & 87.0 ± 0.5  & 94.7 ± 0.2 & 92.8 ± 0.8 & -\\
    \midrule
    Oracle (cumul.) &  -  & 98.6 ± 0.7  & 86.6 ± 1.1  & 82.2 ± 0.3  & 92.1 ± 0.1 & 89.9 ± 0.3 & +1.7 ± 0.4\\
    \bottomrule
  \end{tabular}
\end{table}

\subsubsection{Comparison of Additional Memory Requirement}
\label{sec:extra-mem}
Generally, the regularization-based methods store the old network parameters for regularization or knowledge distillation. In Table \ref{table:memory-requirement}, we compare the extra memory requirement of different regularization-based methods in terms of the number of additional parameters other than the base network parameters. Further, we explain why CLR is the cheapest option from an additional memory requirement perspective compared to other regularization techniques.

In order to quantify the importance of weights to previous tasks, EWC needs to compute and store the diagonal of the Fisher matrix for each task, which has the same number of elements as the network parameters. Additionally, optimal parameters from previous tasks are also stored in order to compute the knowledge distillation loss.  Moreover, a few samples from previous tasks are maintained in our experiments to compute the Fisher matrix after each task is completed. Thus, given that $k$ is the number of encountered tasks and $p$ is the number of network parameters, the space complexity for additional memory usage in EWC becomes $\mathcal{O}(k * p)$. In contrast to EWC, SI computes parameter-specific importance online. So, it does not require storing extra parameters for each task but maintains only the previous task model parameters and regularization strength for each network parameter to compute the surrogate loss. Thus, the space complexity for additional memory becomes $\mathcal{O}(p)$ in the case of SI. MAS also requires storing synaptic importances for each model parameter, needing a similar amount of memory as SI. VCL stores variance term for each weight parameter for current and previous models; hence, it requires thrice the number of additional parameters than the original model. UCL solves this drawback of VCL by computing uncertainty (importance) at node-level, reducing the total required parameters to almost half compared to VCL  \citep{ahn2019uncertainty}. However, CLR outdoes both of them with lesser parameters.

Moreover, both LwF and LFL require storing the old network in the memory to compute the knowledge distillation loss for the previous task. Like SI, these two methods LFL and LwF, have the space complexity for additional memory as $\mathcal{O}(p)$. CLR attempts to achieve a similar objective as the LFL projecting old and new task features in close proximity. However, the CLR is computationally and memory-wise more efficient than the LFL and LwF because the CLR does not need to store the old model and forward pass it, significantly reducing the memory usage and training time.

Hence, all these methods require a large amount of extra memory, which further increases with the number of tasks or the network size. On the other hand, CLR only stores the feature centers of each class, which significantly reduces the additional memory requirement compared to the previous methods. Thus, the space complexity of extra memory usage in CLR comes down to $\mathcal{O}(n * d)$, where $n$ is the number of classes in each task, and each feature center is $d$-dimensional.

\begin{table}[!ht]
\captionsetup{font=small}
\tiny
  \caption{Continual Domain Adaptation (PACS): Test accuracy on individual datasets, average test accuracy (ACC) and BWT  (mean \% and std) after training on all domains for continual domain adaptation with PACS}
  \label{table:domain-adaptation-results-pacs}
\centering
  \begin{tabular}{cccccccc}
    \toprule
    
    Method & M.Size & \bf Sketches (1) & \bf Cartoons (2) & \bf Paintings (3) & \bf Photos (4) & \bf ACC & \bf BWT\\
    \midrule
    \midrule
    Naive  & - & 74.7 ± 2.1 & 65.6 ± 1.6  & 86.7 ± 1.4  & 90.4 ± 0.9 & 77.4 ± 0.5 & -15.0 ± 1.2\\
    \midrule
    LwF   & -  &  79.3 ± 0.8 & 68.4 ± 1.6  & 95.8 ± 1.1  & 88.6 ± 0.5 & 81.4 ± 0.8 & -11.9 ± 1.3\\
    \midrule
    EWC  & -    & 81.1 ± 1.6 & 71.8 ± 0.7  & 96.2 ± 0.4  & 89.1 ± 0.2 & 83.1 ± 0.9 & -8.8 ± 0.8\\
    \midrule
    SI & - & 82.6 ± 1.1 & 70.0 ± 1.4  & 92.2 ± 1.6  & 90.5 ± 0.7 & 83.7 ± 1.2 & -10.2 ± 0.8\\
    \midrule
    LFL &  -  & 86.6 ± 0.7 & 73.5 ± 0.9  & 95.2 ± 1.2  & 90.4 ± 1.4 & 85.7 ± 1.1 & -7.5 ± 1.0 \\
    \midrule
    Meta-DR &  - & 86.8 ± 1.2 & 75.4 ± 0.7 & 95.3 ± 0.9 & 88.5 ± 0.3 & 85.9 ± 0.9 & -7.2 ± 0.8 \\
    \midrule
     CLR  &  -  & 86.7 ± 1.4 & 74.0 ± 0.8  & 97.0 ± 1.2  & 89.9 ± 0.7 & 86.1 ± 1.3 & -7.9 ± 0.4\\
    \midrule
    \midrule
    & 10  & 89.2 ± 1.2 & 84.3 ± 0.3  & 96.8 ± 0.9  & 91.8 ± 1.9 & 89.9 ± 1.5 & -3.4 ± 0.6\\
      ER & 20  & 92.3 ± 0.8 & 83.1 ± 1.2  & 96.4 ± 2.1  & 90.7 ± 0.7 & 90.1 ± 1.7  & -2.0 ± 1.3\\
      & 30 & 91.7 ± 0.5 & 84.8 ± 1.6  & 97.6 ± 2.4  & 91.5 ± 0.4 & 91.0 ± 1.2  & -2.2 ± 1.1\\
      \midrule
    \midrule
    & 10  & 87.2 ± 1.1 & 85.0 ± 0.8  & 97.1 ± 0.9  & 90.2 ± 1.9 & 89.8 ± 0.9 & -4.1 ± 0.8\\
      A-GEM & 20  & 91.8 ± 0.7 & 84.1 ± 0.8  & 95.4 ± 1.7  & 91.1 ± 0.7 & 90.6 ± 1.4  & -2.7 ± 1.2\\
      & 30 & 91.2 ± 1.2 & 84.5 ± 0.9  & 97.1 ± 1.4  & 90.3 ± 2.2 & 90.7 ± 1.2  & -2.4 ± 1.4\\
      \midrule
    \midrule
    & 10  & 90.2 ± 0.4 & 84.3 ± 1.7  & 95.1 ± 1.5  & 90.3 ± 1.9 & 90.0 ± 1.5 & -3.5 ± 1.9\\
      ER + Meta-DR & 20  & 92.5 ± 1.8 & 82.9 ± 2.2  & 95.6 ± 1.2  & 91.2 ± 1.7 & 90.5 ± 0.6  & -2.1 ± 1.1\\
      & 30 & 90.7 ± 1.6 & 84.9 ± 0.8  & 97.8 ± 2.1  & 89.4 ± 1.4 & 90.7 ± 1.2  & -2.6 ± 1.3\\
    \midrule
    \midrule
    & 10 & 90.7 ± 0.5 & 84.1 ± 0.7  & 97.4 ± 1.4  & 91.5 ± 0.6  & 90.4 ± 1.1  & -3.6 ± 0.7\\
    
     ER + CLR & 20 & 91.5 ± 0.6 & 85.4 ± 1.6  & 96.6 ± 1.4  & 91.4 ± 0.8 & 90.9 ± 1.3  & -2.8 ± 0.5\\
    
     & 30 & 91.3 ± 1.0 & 87.1 ± 0.9  & 96.8 ± 0.4  & 92.4 ± 0.6 & 91.4 ± 0.8  & -2.9 ± 1.0\\
    \midrule
    \midrule
    Oracle (all) & - & 93.5 ± 1.5 & 91.3 ± 1.8  & 95.0 ± 0.7  & 85.6 ± 1.2 & 91.6 ± 0.8 & -\\
    \midrule
    Oracle (cumul.) & - & 94.3 ± 2.3 & 92.3 ± 1.9  & 96.2 ± 1.4  & 91.0 ± 0.9 & 93.5 ± 1.6 & +0.8 ± 0.5\\
    \bottomrule
  \end{tabular}
\end{table}

\begin{table}[!ht]
\captionsetup{font=small}
\tiny
  \caption{Comparison of additional memory requirement for all methods}
  \label{table:memory-requirement}
\centering
  \begin{tabular}{lllllllll}
    \toprule
    Protocol/Method      & 
    LwF & EWC & SI & LFL & MAS & UCL & VCL & CLR\\
    \midrule
    \midrule
    MNIST (MLP)  & 89.6K & 1.79M & 179K & 89.6K & 179K & 89.8K & 269K & 1000\\
    \midrule
    Digits, PACS (ResNet18) & 11M & 88M & 22M & 11M & 22M & 11M & 33M & 5120, 3584 \\
    \bottomrule
  \end{tabular}
\end{table}

\section{Limitations and Future Directions}
\label{sec:limitations}

There are several exciting research directions to extend our work for continual learning. Our method requires the knowledge of task boundaries which may not always be available. CLR does not leverage the task descriptors, which may be exploited to obtain a positive forward transfer. Further, novel approaches can also be developed to exploit our approach for supporting task-incremental and class-incremental learning, where task-IDs need to be inferred. In this paper, we applied CLR to solve the supervised classification problem in continual learning. It would be an interesting research direction to exploit the properties of CLR to solve other problems like regression and dimensionality reduction in the continual learning setting. Moreover, the effects of using other discriminative representation learning approaches \citep{hadsell2006dimensionality, sun2015deep, deng2017marginal, zhang2017range, liu2017sphereface, chen2017noisy, wan2018rethinking, qi2018face, wang2018additive, wang2018cosface} can be studied for continual learning.

\section{Conclusion}
\label{sec:conclusion}

In this paper, we proposed a new regularization-based strategy for continual learning, referred to as \textit{center loss regularization (CLR)}. It utilizes the power of center loss to learn discriminative features and use the learned feature centers to project new task features in the proximity of old task features to transfer knowledge and avoid catastrophic forgetting. Our method was effective in overcoming catastrophic forgetting when applied to the standard continual learning benchmarks as well as continual domain adaptation benchmarks. Our method is scalable and computationally effective, and it does not store previous data and requires minimal additional network parameters. Our extensive experiments consistently demonstrate the competitive performance of CLR against the state-of-the-art regularization strategies for continual learning.

\bibliography{iclr2022_conference}

\begin{thebibliography}{63}
\providecommand{\natexlab}[1]{#1}
\providecommand{\url}[1]{\texttt{#1}}
\expandafter\ifx\csname urlstyle\endcsname\relax
  \providecommand{\doi}[1]{doi: #1}\else
  \providecommand{\doi}{doi: \begingroup \urlstyle{rm}\Url}\fi

\bibitem[Adel et~al.(2020)Adel, Zhao, and Turner]{Adel2020Continual}
Tameem Adel, Han Zhao, and Richard~E. Turner.
\newblock Continual learning with adaptive weights (claw).
\newblock In \emph{International Conference on Learning Representations}, 2020.

\bibitem[Ahn et~al.(2019)Ahn, Cha, Lee, and Moon]{ahn2019uncertainty}
Hongjoon Ahn, Sungmin Cha, Donggyu Lee, and Taesup Moon.
\newblock Uncertainty-based continual learning with adaptive regularization.
\newblock In \emph{Advances in Neural Information Processing Systems}, pp.\
  4394--4404, 2019.

\bibitem[Aljundi et~al.(2018)Aljundi, Babiloni, Elhoseiny, Rohrbach, and
  Tuytelaars]{aljundi2018memory}
Rahaf Aljundi, Francesca Babiloni, Mohamed Elhoseiny, Marcus Rohrbach, and
  Tinne Tuytelaars.
\newblock Memory aware synapses: Learning what (not) to forget.
\newblock In \emph{Proceedings of the European Conference on Computer Vision
  (ECCV)}, pp.\  139--154, 2018.

\bibitem[Chaudhry et~al.(2019)Chaudhry, Ranzato, Rohrbach, and Elhoseiny]{AGEM}
Arslan Chaudhry, Marc’Aurelio Ranzato, Marcus Rohrbach, and Mohamed
  Elhoseiny.
\newblock Efficient lifelong learning with a-gem.
\newblock In \emph{ICLR}, 2019.

\bibitem[Chen et~al.(2017)Chen, Deng, and Du]{chen2017noisy}
B.~Chen, W.~Deng, and J.~Du.
\newblock Noisy softmax: Improving the generalization ability of dcnn via
  postponing the early softmax saturation.
\newblock In \emph{2017 IEEE Conference on Computer Vision and Pattern
  Recognition (CVPR)}, pp.\  4021--4030, 2017.

\bibitem[Chen \& Liu(2018)Chen and Liu]{chen2018lifelong}
Zhiyuan Chen and Bing Liu.
\newblock Lifelong machine learning.
\newblock \emph{Synthesis Lectures on Artificial Intelligence and Machine
  Learning}, 12\penalty0 (3):\penalty0 1--207, 2018.

\bibitem[Deng et~al.(2009)Deng, Dong, Socher, Li, Li, and
  Fei-Fei]{deng2009imagenet}
Jia Deng, Wei Dong, Richard Socher, Li-Jia Li, Kai Li, and Li~Fei-Fei.
\newblock Imagenet: A large-scale hierarchical image database.
\newblock In \emph{2009 IEEE conference on computer vision and pattern
  recognition}, pp.\  248--255. Ieee, 2009.

\bibitem[Deng et~al.(2017)Deng, Zhou, and Zafeiriou]{deng2017marginal}
Jiankang Deng, Yuxiang Zhou, and Stefanos Zafeiriou.
\newblock Marginal loss for deep face recognition.
\newblock In \emph{2017 IEEE Conference on Computer Vision and Pattern
  Recognition Workshops (CVPRW)}, pp.\  2006--2014, 2017.

\bibitem[Deng et~al.(2019)Deng, Guo, Xue, and Zafeiriou]{deng2019arcface}
Jiankang Deng, Jia Guo, Niannan Xue, and Stefanos Zafeiriou.
\newblock Arcface: Additive angular margin loss for deep face recognition.
\newblock In \emph{2019 IEEE/CVF Conference on Computer Vision and Pattern
  Recognition (CVPR)}, pp.\  4685--4694, 2019.

\bibitem[Farabet et~al.(2013)Farabet, Couprie, Najman, and LeCun]{6338939}
Clement Farabet, Camille Couprie, Laurent Najman, and Yann LeCun.
\newblock Learning hierarchical features for scene labeling.
\newblock \emph{IEEE Transactions on Pattern Analysis and Machine
  Intelligence}, 35\penalty0 (8):\penalty0 1915--1929, 2013.

\bibitem[Ganin \& Lempitsky(2015)Ganin and Lempitsky]{ganin2015unsupervised}
Yaroslav Ganin and Victor Lempitsky.
\newblock Unsupervised domain adaptation by backpropagation.
\newblock In \emph{International conference on machine learning}, pp.\
  1180--1189. PMLR, 2015.

\bibitem[Grossberg(1982)]{grossberg1982does}
Stephen Grossberg.
\newblock How does a brain build a cognitive code?
\newblock \emph{Studies of mind and brain}, pp.\  1--52, 1982.

\bibitem[Grossberg(2013)]{grossberg2013adaptive}
Stephen Grossberg.
\newblock Adaptive resonance theory: How a brain learns to consciously attend,
  learn, and recognize a changing world.
\newblock \emph{Neural networks}, 37:\penalty0 1--47, 2013.

\bibitem[Guest \& Love(2019)Guest and Love]{Guest626374}
Olivia Guest and Bradley~C. Love.
\newblock Levels of representation in a deep learning model of categorization.
\newblock \emph{bioRxiv}, 2019.
\newblock \doi{10.1101/626374}.

\bibitem[Hadsell et~al.(2006)Hadsell, Chopra, and
  LeCun]{hadsell2006dimensionality}
Raia Hadsell, Sumit Chopra, and Yann LeCun.
\newblock Dimensionality reduction by learning an invariant mapping.
\newblock In \emph{2006 IEEE Computer Society Conference on Computer Vision and
  Pattern Recognition (CVPR'06)}, volume~2, pp.\  1735--1742. IEEE, 2006.

\bibitem[He et~al.(2016)He, Zhang, Ren, and Sun]{he2016deep}
Kaiming He, Xiangyu Zhang, Shaoqing Ren, and Jian Sun.
\newblock Deep residual learning for image recognition.
\newblock In \emph{2016 IEEE conference on computer vision and pattern
  recognition}, pp.\  770--778, 2016.

\bibitem[Hecht-Nielsen(1992)]{hecht1992theory}
Robert Hecht-Nielsen.
\newblock Theory of the backpropagation neural network.
\newblock In \emph{Neural networks for perception}, pp.\  65--93. Elsevier,
  1992.

\bibitem[Jung et~al.(2016)Jung, Ju, Jung, and Kim]{Jung2016LessforgettingLI}
Heechul Jung, Jeongwoo Ju, Minju Jung, and Junmo Kim.
\newblock Less-forgetting learning in deep neural networks.
\newblock \emph{ArXiv}, abs/1607.00122, 2016.

\bibitem[Jung et~al.(2018)Jung, Ju, Jung, and Kim]{jung2018less}
Heechul Jung, Jeongwoo Ju, Minju Jung, and Junmo Kim.
\newblock Less-forgetful learning for domain expansion in deep neural networks.
\newblock In \emph{Thirty-Second AAAI Conference on Artificial Intelligence},
  2018.

\bibitem[Kingma \& Ba(2014)Kingma and Ba]{kingma2014adam}
Diederik~P Kingma and Jimmy Ba.
\newblock Adam: A method for stochastic optimization.
\newblock \emph{arXiv preprint arXiv:1412.6980}, 2014.

\bibitem[Kirkpatrick et~al.(2017)Kirkpatrick, Pascanu, Rabinowitz, Veness,
  Desjardins, Rusu, Milan, Quan, Ramalho, Grabska-Barwinska,
  et~al.]{kirkpatrick2017overcoming}
James Kirkpatrick, Razvan Pascanu, Neil Rabinowitz, Joel Veness, Guillaume
  Desjardins, Andrei~A Rusu, Kieran Milan, John Quan, Tiago Ramalho, Agnieszka
  Grabska-Barwinska, et~al.
\newblock Overcoming catastrophic forgetting in neural networks.
\newblock \emph{Proceedings of the national academy of sciences}, 114\penalty0
  (13):\penalty0 3521--3526, 2017.

\bibitem[Koch et~al.(2015)Koch, Zemel, and Salakhutdinov]{koch2015siamese}
Gregory Koch, Richard Zemel, and Ruslan Salakhutdinov.
\newblock Siamese neural networks for one-shot image recognition.
\newblock In \emph{ICML deep learning workshop}, volume~2. Lille, 2015.

\bibitem[LeCun(1998)]{lecun1998mnist}
Yann LeCun.
\newblock The mnist database of handwritten digits.
\newblock \emph{http://yann. lecun. com/exdb/mnist/}, 1998.

\bibitem[Li et~al.(2017)Li, Yang, Song, and Hospedales]{li2017deeper}
Da~Li, Yongxin Yang, Yi-Zhe Song, and Timothy~M Hospedales.
\newblock Deeper, broader and artier domain generalization.
\newblock In \emph{Proceedings of the IEEE international conference on computer
  vision}, pp.\  5542--5550, 2017.

\bibitem[Li \& Hoiem(2017)Li and Hoiem]{li2017learning}
Zhizhong Li and Derek Hoiem.
\newblock Learning without forgetting.
\newblock \emph{IEEE transactions on pattern analysis and machine
  intelligence}, 40\penalty0 (12):\penalty0 2935--2947, 2017.

\bibitem[Liu et~al.(2016)Liu, Wen, Yu, and Yang]{liu2016large}
Weiyang Liu, Yandong Wen, Zhiding Yu, and Meng Yang.
\newblock Large-margin softmax loss for convolutional neural networks.
\newblock In \emph{ICML}, volume~2, pp.\ ~7, 2016.

\bibitem[Liu et~al.(2017)Liu, Wen, Yu, Li, Raj, and Song]{liu2017sphereface}
Weiyang Liu, Yandong Wen, Zhiding Yu, Ming Li, Bhiksha Raj, and Le~Song.
\newblock Sphereface: Deep hypersphere embedding for face recognition.
\newblock In \emph{2017 IEEE conference on computer vision and pattern
  recognition}, pp.\  212--220, 2017.

\bibitem[Lomonaco \& Maltoni(2017)Lomonaco and Maltoni]{lomonaco2017core50}
Vincenzo Lomonaco and Davide Maltoni.
\newblock Core50: a new dataset and benchmark for continuous object
  recognition.
\newblock In \emph{Conference on Robot Learning}, pp.\  17--26. PMLR, 2017.

\bibitem[Lomonaco et~al.(2021)Lomonaco, Pellegrini, Cossu, Carta, Graffieti,
  Hayes, De~Lange, Masana, Pomponi, van~de Ven, Mundt, She, Cooper, Forest,
  Belouadah, Calderara, Parisi, Cuzzolin, Tolias, Scardapane, Antiga, Ahmad,
  Popescu, Kanan, van~de Weijer, Tuytelaars, Bacciu, and
  Maltoni]{lomonaco2021avalanche}
Vincenzo Lomonaco, Lorenzo Pellegrini, Andrea Cossu, Antonio Carta, Gabriele
  Graffieti, Tyler~L. Hayes, Matthias De~Lange, Marc Masana, Jary Pomponi,
  Gido~M. van~de Ven, Martin Mundt, Qi~She, Keiland Cooper, Jeremy Forest, Eden
  Belouadah, Simone Calderara, German~I. Parisi, Fabio Cuzzolin, Andreas~S.
  Tolias, Simone Scardapane, Luca Antiga, Subutai Ahmad, Adrian Popescu,
  Christopher Kanan, Joost van~de Weijer, Tinne Tuytelaars, Davide Bacciu, and
  Davide Maltoni.
\newblock Avalanche: an end-to-end library for continual learning.
\newblock In \emph{2021 IEEE/CVF Conference on Computer Vision and Pattern
  Recognition Workshops (CVPRW)}, pp.\  3595--3605, 2021.

\bibitem[Lopez-Paz \& Ranzato(2017)Lopez-Paz and Ranzato]{lopez2017gradient}
David Lopez-Paz and Marc\textquotesingle~Aurelio Ranzato.
\newblock Gradient episodic memory for continual learning.
\newblock In I.~Guyon, U.~V. Luxburg, S.~Bengio, H.~Wallach, R.~Fergus,
  S.~Vishwanathan, and R.~Garnett (eds.), \emph{Advances in Neural Information
  Processing Systems}, volume~30, 2017.

\bibitem[McCloskey \& Cohen(1989)McCloskey and
  Cohen]{mccloskey1989catastrophic}
Michael McCloskey and Neal~J Cohen.
\newblock Catastrophic interference in connectionist networks: The sequential
  learning problem.
\newblock In \emph{Psychology of learning and motivation}, volume~24, pp.\
  109--165. Elsevier, 1989.

\bibitem[Mermillod et~al.(2013)Mermillod, Bugaiska, and
  Bonin]{mermillod2013stability}
Martial Mermillod, Aur{\'e}lia Bugaiska, and Patrick Bonin.
\newblock The stability-plasticity dilemma: Investigating the continuum from
  catastrophic forgetting to age-limited learning effects.
\newblock \emph{Frontiers in psychology}, 4:\penalty0 504, 2013.

\bibitem[Netzer et~al.(2011)Netzer, Wang, Coates, Bissacco, Wu, and
  Ng]{netzer2011reading}
Yuval Netzer, Tao Wang, Adam Coates, Alessandro Bissacco, Bo~Wu, and Andrew Ng.
\newblock Reading digits in natural images with unsupervised feature learning.
\newblock \emph{NIPS}, 01 2011.

\bibitem[Nguyen et~al.(2017)Nguyen, Li, Bui, and Turner]{nguyen2017variational}
Cuong~V Nguyen, Yingzhen Li, Thang~D Bui, and Richard~E Turner.
\newblock Variational continual learning.
\newblock \emph{arXiv preprint arXiv:1710.10628}, 2017.

\bibitem[Parisi et~al.(2019)Parisi, Kemker, Part, Kanan, and
  Wermter]{parisi2019continual}
German~I Parisi, Ronald Kemker, Jose~L Part, Christopher Kanan, and Stefan
  Wermter.
\newblock Continual lifelong learning with neural networks: A review.
\newblock \emph{Neural Networks}, 113:\penalty0 54--71, 2019.

\bibitem[Pellegrini et~al.(2020)Pellegrini, Graffieti, Lomonaco, and
  Maltoni]{pellegrini2020latent}
Lorenzo Pellegrini, Gabriele Graffieti, Vincenzo Lomonaco, and Davide Maltoni.
\newblock Latent replay for real-time continual learning.
\newblock In \emph{2020 IEEE/RSJ International Conference on Intelligent Robots
  and Systems (IROS)}, pp.\  10203--10209. IEEE, 2020.

\bibitem[Prabhu et~al.(2020)Prabhu, Torr, and Dokania]{prabhu2020greedy}
Ameya Prabhu, Philip Torr, and Puneet Dokania.
\newblock Gdumb: A simple approach that questions our progress in continual
  learning.
\newblock In \emph{The European Conference on Computer Vision (ECCV)}, August
  2020.

\bibitem[Qi \& Zhang(2018)Qi and Zhang]{qi2018face}
Xianbiao Qi and Lei Zhang.
\newblock Face recognition via centralized coordinate learning.
\newblock \emph{arXiv preprint arXiv:1801.05678}, 2018.

\bibitem[Ramasesh et~al.(2020)Ramasesh, Dyer, and Raghu]{ramasesh2020anatomy}
Vinay~V Ramasesh, Ethan Dyer, and Maithra Raghu.
\newblock Anatomy of catastrophic forgetting: Hidden representations and task
  semantics.
\newblock \emph{arXiv preprint arXiv:2007.07400}, 2020.

\bibitem[Ramirez et~al.(2013)Ramirez, Liu, Lin, Suh, Pignatelli, Redondo, Ryan,
  and Tonegawa]{ramirez2013creating}
Steve Ramirez, Xu~Liu, Pei-Ann Lin, Junghyup Suh, Michele Pignatelli, Roger~L
  Redondo, Tom{\'a}s~J Ryan, and Susumu Tonegawa.
\newblock Creating a false memory in the hippocampus.
\newblock \emph{Science}, 341\penalty0 (6144):\penalty0 387--391, 2013.

\bibitem[Ratcliff(1990)]{ratcliff1990connectionist}
Roger Ratcliff.
\newblock Connectionist models of recognition memory: constraints imposed by
  learning and forgetting functions.
\newblock \emph{Psychological review}, 97\penalty0 (2):\penalty0 285, 1990.

\bibitem[Richardson \& Thomas(2008)Richardson and
  Thomas]{richardson2008critical}
Fiona~M Richardson and Michael~SC Thomas.
\newblock Critical periods and catastrophic interference effects in the
  development of self-organizing feature maps.
\newblock \emph{Developmental science}, 11\penalty0 (3):\penalty0 371--389,
  2008.

\bibitem[Rolnick et~al.(2018)Rolnick, Ahuja, Schwarz, Lillicrap, and
  Wayne]{rolnick2018experience}
David Rolnick, Arun Ahuja, Jonathan Schwarz, Timothy~P Lillicrap, and Greg
  Wayne.
\newblock Experience replay for continual learning.
\newblock \emph{arXiv preprint arXiv:1811.11682}, 2018.

\bibitem[Rusu et~al.(2016)Rusu, Rabinowitz, Desjardins, Soyer, Kirkpatrick,
  Kavukcuoglu, Pascanu, and Hadsell]{rusu2016progressive}
Andrei~A Rusu, Neil~C Rabinowitz, Guillaume Desjardins, Hubert Soyer, James
  Kirkpatrick, Koray Kavukcuoglu, Razvan Pascanu, and Raia Hadsell.
\newblock Progressive neural networks.
\newblock \emph{arXiv preprint arXiv:1606.04671}, 2016.

\bibitem[Schroff et~al.(2015)Schroff, Kalenichenko, and
  Philbin]{schroff2015facenet}
Florian Schroff, Dmitry Kalenichenko, and James Philbin.
\newblock Facenet: A unified embedding for face recognition and clustering.
\newblock In \emph{2015 IEEE conference on computer vision and pattern
  recognition}, pp.\  815--823, 2015.

\bibitem[Shin et~al.(2017)Shin, Lee, Kim, and Kim]{shin2017continual}
Hanul Shin, Jung~Kwon Lee, Jaehong Kim, and Jiwon Kim.
\newblock Continual learning with deep generative replay.
\newblock \emph{arXiv preprint arXiv:1705.08690}, 2017.

\bibitem[Stickgold \& Walker(2007)Stickgold and Walker]{stickgold2007sleep}
Robert Stickgold and Matthew~P Walker.
\newblock Sleep-dependent memory consolidation and reconsolidation.
\newblock \emph{Sleep medicine}, 8\penalty0 (4):\penalty0 331--343, 2007.

\bibitem[Sun(2015)]{sun2015deep}
Yi~Sun.
\newblock \emph{Deep learning face representation by joint
  identification-verification}.
\newblock The Chinese University of Hong Kong (Hong Kong), 2015.

\bibitem[Sussmann(1992)]{sussmann1992uniqueness}
H{\'e}ctor~J Sussmann.
\newblock Uniqueness of the weights for minimal feedforward nets with a given
  input-output map.
\newblock \emph{Neural networks}, 5\penalty0 (4):\penalty0 589--593, 1992.

\bibitem[Thrun(1995)]{thrun1995lifelong}
Sebastian Thrun.
\newblock A lifelong learning perspective for mobile robot control.
\newblock In \emph{Intelligent robots and systems}, pp.\  201--214. Elsevier,
  1995.

\bibitem[Thrun(1996)]{thrun1996learning}
Sebastian Thrun.
\newblock Is learning the n-th thing any easier than learning the first?
\newblock In \emph{Advances in neural information processing systems}, pp.\
  640--646. Morgan Kaufmann Publishers, 1996.

\bibitem[Thrun(1998)]{thrun1998lifelong}
Sebastian Thrun.
\newblock Lifelong learning algorithms.
\newblock In \emph{Learning to learn}, pp.\  181--209. Springer, 1998.

\bibitem[Thrun \& Pratt(2012)Thrun and Pratt]{thrun2012learning}
Sebastian Thrun and Lorien Pratt.
\newblock \emph{Learning to learn}.
\newblock Springer Science \& Business Media, 2012.

\bibitem[Van~de Ven \& Tolias(2018)Van~de Ven and Tolias]{van2018generative}
Gido~M Van~de Ven and Andreas~S Tolias.
\newblock Generative replay with feedback connections as a general strategy for
  continual learning.
\newblock \emph{arXiv preprint arXiv:1809.10635}, 2018.

\bibitem[Van~de Ven \& Tolias(2019)Van~de Ven and Tolias]{van2019three}
Gido~M Van~de Ven and Andreas~S Tolias.
\newblock Three scenarios for continual learning.
\newblock \emph{arXiv preprint arXiv:1904.07734}, 2019.

\bibitem[van~de Ven et~al.(2020)van~de Ven, Siegelmann, and
  Tolias]{van2020brain}
Gido~M van~de Ven, Hava~T Siegelmann, and Andreas~S Tolias.
\newblock Brain-inspired replay for continual learning with artificial neural
  networks.
\newblock \emph{Nature communications}, 11\penalty0 (1):\penalty0 1--14, 2020.

\bibitem[Volpi et~al.(2021)Volpi, Larlus, and Rogez]{Volpi2021ContinualAO}
Riccardo Volpi, Diane Larlus, and Gr{\'e}gory Rogez.
\newblock Continual adaptation of visual representations via domain
  randomization and meta-learning.
\newblock In \emph{Proceedings of the IEEE/CVF Conference on Computer Vision
  and Pattern Recognition}, pp.\  4443--4453, 2021.

\bibitem[Wan et~al.(2018)Wan, Zhong, Li, and Chen]{wan2018rethinking}
Weitao Wan, Yuanyi Zhong, Tianpeng Li, and Jiansheng Chen.
\newblock Rethinking feature distribution for loss functions in image
  classification.
\newblock In \emph{2018 IEEE conference on computer vision and pattern
  recognition}, pp.\  9117--9126, 2018.

\bibitem[Wang et~al.(2018{\natexlab{a}})Wang, Cheng, Liu, and
  Liu]{wang2018additive}
Feng Wang, Jian Cheng, Weiyang Liu, and Haijun Liu.
\newblock Additive margin softmax for face verification.
\newblock \emph{IEEE Signal Processing Letters}, 25\penalty0 (7):\penalty0
  926--930, 2018{\natexlab{a}}.

\bibitem[Wang et~al.(2018{\natexlab{b}})Wang, Wang, Zhou, Ji, Gong, Zhou, Li,
  and Liu]{wang2018cosface}
Hao Wang, Yitong Wang, Zheng Zhou, Xing Ji, Dihong Gong, Jingchao Zhou, Zhifeng
  Li, and Wei Liu.
\newblock Cosface: Large margin cosine loss for deep face recognition.
\newblock In \emph{2018 IEEE conference on computer vision and pattern
  recognition}, pp.\  5265--5274, 2018{\natexlab{b}}.

\bibitem[Wen et~al.(2016)Wen, Zhang, Li, and Qiao]{wen2016discriminative}
Yandong Wen, Kaipeng Zhang, Zhifeng Li, and Yu~Qiao.
\newblock A discriminative feature learning approach for deep face recognition.
\newblock In \emph{European conference on computer vision}, pp.\  499--515.
  Springer, 2016.

\bibitem[Zenke et~al.(2017)Zenke, Poole, and Ganguli]{zenke2017continual}
Friedemann Zenke, Ben Poole, and Surya Ganguli.
\newblock Continual learning through synaptic intelligence.
\newblock In \emph{International Conference on Machine Learning}, pp.\
  3987--3995. PMLR, 2017.

\bibitem[Zhang et~al.(2017)Zhang, Fang, Wen, Li, and Qiao]{zhang2017range}
Xiao Zhang, Zhiyuan Fang, Yandong Wen, Zhifeng Li, and Yu~Qiao.
\newblock Range loss for deep face recognition with long-tailed training data.
\newblock In \emph{Proceedings of the IEEE International Conference on Computer
  Vision}, pp.\  5409--5418, 2017.

\end{thebibliography}
\bibliographystyle{iclr2022_conference}

\appendix

\section{Training Details}
\label{appendix:Appendix A}

We used one NVIDIA GeForce TITAN X GPU for training the models in all our experiments with the CUDA version 10.1 on a Linux machine. We built our models and implemented different methods using PyTorch deep learning framework and Avalanche library \citep{lomonaco2021avalanche} for Continual Learning. Hyperparameter search was done using the grid-search. Next, we provide the best hyperparameter values, specific to the methods for our experiments. The \(\lambda\) denotes the importance of the regularization penalty for EWC, LwF, SI, LFL and MAS. Whereas $\beta$ controls the speed of standard deviation ($\sigma$) for the weight parameter in UCL. VCL does not need any additional hyperparameters. In CLR, \(\lambda\) denotes the importance of the center loss and \(\alpha\) denotes the rate with which the centers are allowed to change in the first task.

\begin{table}[ht]
\small
  \caption{Hyperparameters specific to methods in all experiments}
  \label{table:hyper-params}
  \begin{center}
  \begin{tabular}{ccccc}
    \toprule
    
   \bf  Method & \bf Rotated MNIST & \bf Permuted MNIST & \bf Digits & \bf PACS \\
    \midrule
    \midrule
    EWC ($\lambda$)  & 0.001 & 0.001  & 0.001  &  1.0 \\
    \midrule
    LwF ($\lambda$)  & 1.0 & 1.0  & 0.1  & 0.1 \\
    \midrule
    SI ($\lambda$)  & 0.001 & 0.001  & 0.1  & 1.0 \\
    \midrule
    LFL ($\lambda$)  & 1.0 & 0.001  & 0.1  & 1.0 \\
    \midrule
    MAS ($\lambda$)  & 10.0 & 1.0  & -  & - \\
    \midrule
    UCL ($\beta$)  & 0.9 & 1.0  & -  & - \\
    \midrule
    CLR (\(\lambda\), \(\alpha\)) & (0.001, 0.03) & (0.04, 0.03)  & (0.01, 0.01)  & (0.001, 0.5) \\
    \bottomrule
  \end{tabular}
  \end{center}
 \end{table}

\section{Evaluation Metrics}
\label{appendix:Appendix B}

\citet{lopez2017gradient} introduced Average Accuracy (ACC) and Backward Transfer (BWT) evaluation metrics for continual learning, which we use in our experiments to evaluate our approach. For evaluation, we maintain a test set for each of the $T$ tasks. After learning on the new task \(t_i\), we evaluate the model's test performance on all $T$ tasks. The formulas for calculating ACC and BWT are as follows. $A_{i,j}$ is the test classification accuracy of the model on task $t_i$ after observing the last data
sample from task $t_j$.
\begin{align} 
\label{eq:acc}
\begin{split}
\mathbf{Average\ Accuracy\ (ACC)} & = \frac{1}{T} \sum_{i=1}^{T} A_{i,T}
\end{split}
\\
\label{eq:bwt}
\begin{split}
\mathbf{Backward\ Transfer\ (BWT)} & = \frac{1}{T-1} \sum_{i=1}^{T-1} A_{i, T} - A_{i, i}
\end{split}
\end{align}

The larger the value of ACC, the better is the model. Whereas, if the values of ACC of two models are similar, then the model with higher BWT is usually considered.

\section{Additional Plots}
\label{appendix:additional_plots}
\begin{figure}[H]
    \centering
    \begin{minipage}{0.48\textwidth}
        \includegraphics[width=1.0\textwidth]{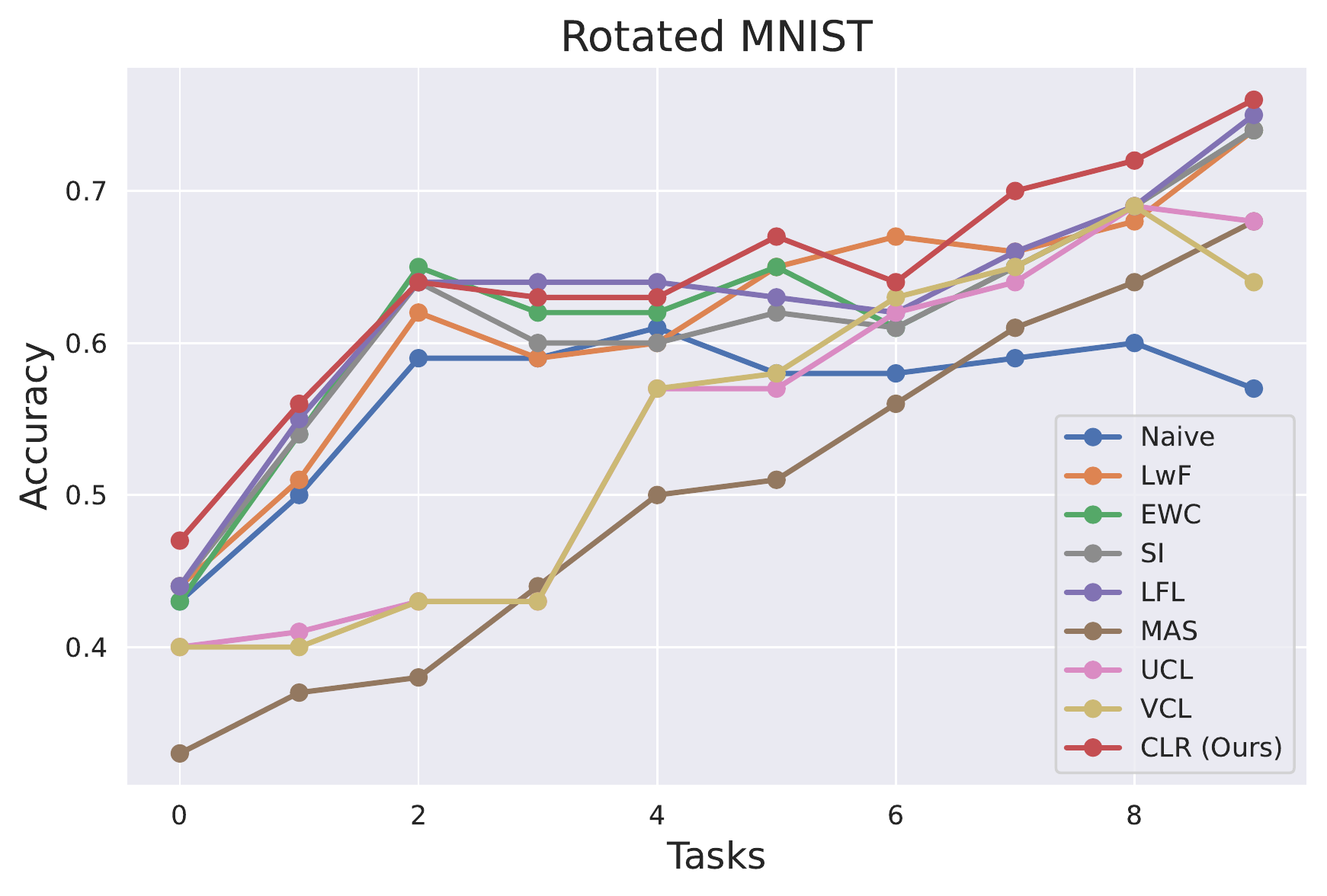}
    \end{minipage}
    \vspace{3mm}
    \centering
    \begin{minipage}{0.48\textwidth}
        \centering
        \includegraphics[width=1.0\textwidth]{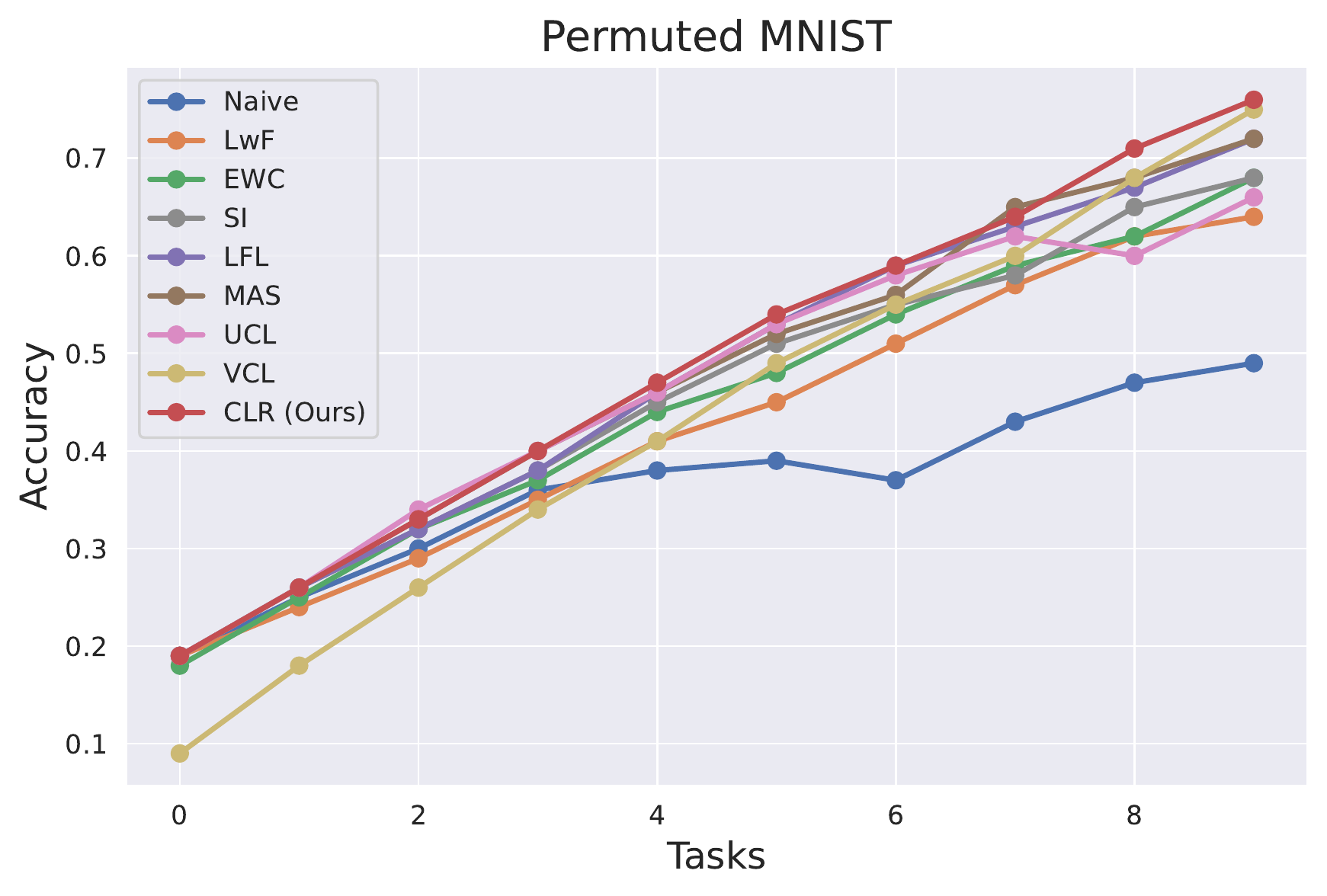} 
    \end{minipage}
    
\caption{Average test accuracy on all T tasks after training on each task for all methods on Rotated and Permuted MNIST}
\label{fig:appendix-plots}
\end{figure}

\begin{figure}[ht]
    \centering
    \begin{minipage}{0.6\textwidth}
        \includegraphics[width=1.0\textwidth]{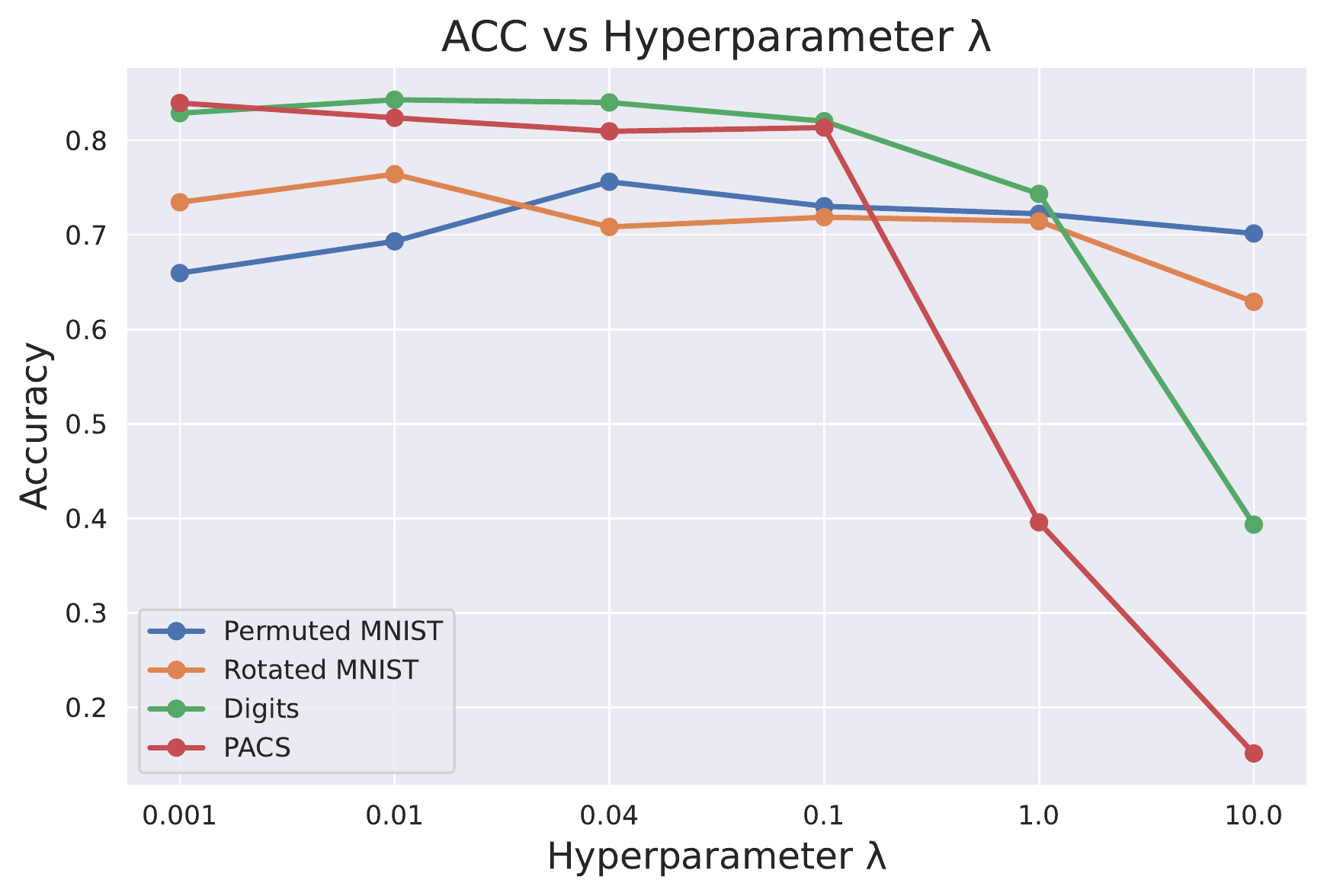}
    \end{minipage}
    
\caption{Average test accuracy (ACC) for various values of CLR's hyperparameter $\lambda$ for different protocols}
\label{fig:lambda-plots}
\end{figure}

Figure \ref{fig:appendix-plots} shows the plots of average accuracy (over 10 tasks) at the end of training on each task. This is different from the plots presented in Figure \ref{fig:results-mnist} where we take the average over encountered tasks only. In Figure \ref{fig:lambda-plots}, we show the performance of CLR for various values of hyperparameter $\lambda$ for various protocols. The value of $\lambda$ at the highest/peak point of each protocol's graph was chosen for the corresponding experiment.

\begin{figure}[!ht]
    \centering
    \begin{minipage}{0.48\textwidth}
        \includegraphics[width=1.0\textwidth]{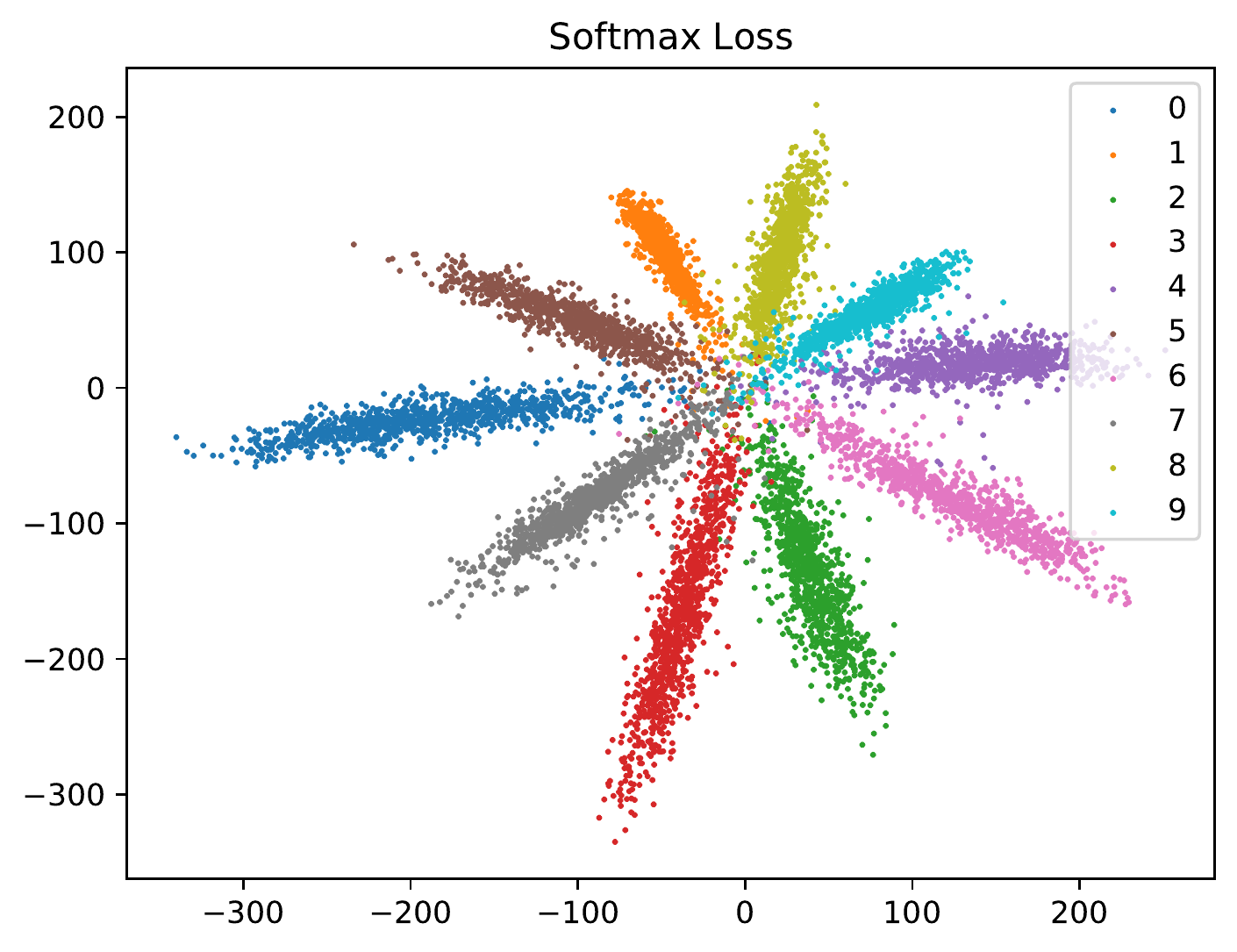}
    \end{minipage}
    \vspace{3mm}
    \centering
    \begin{minipage}{0.48\textwidth}
        \centering
        \includegraphics[width=1.0\textwidth]{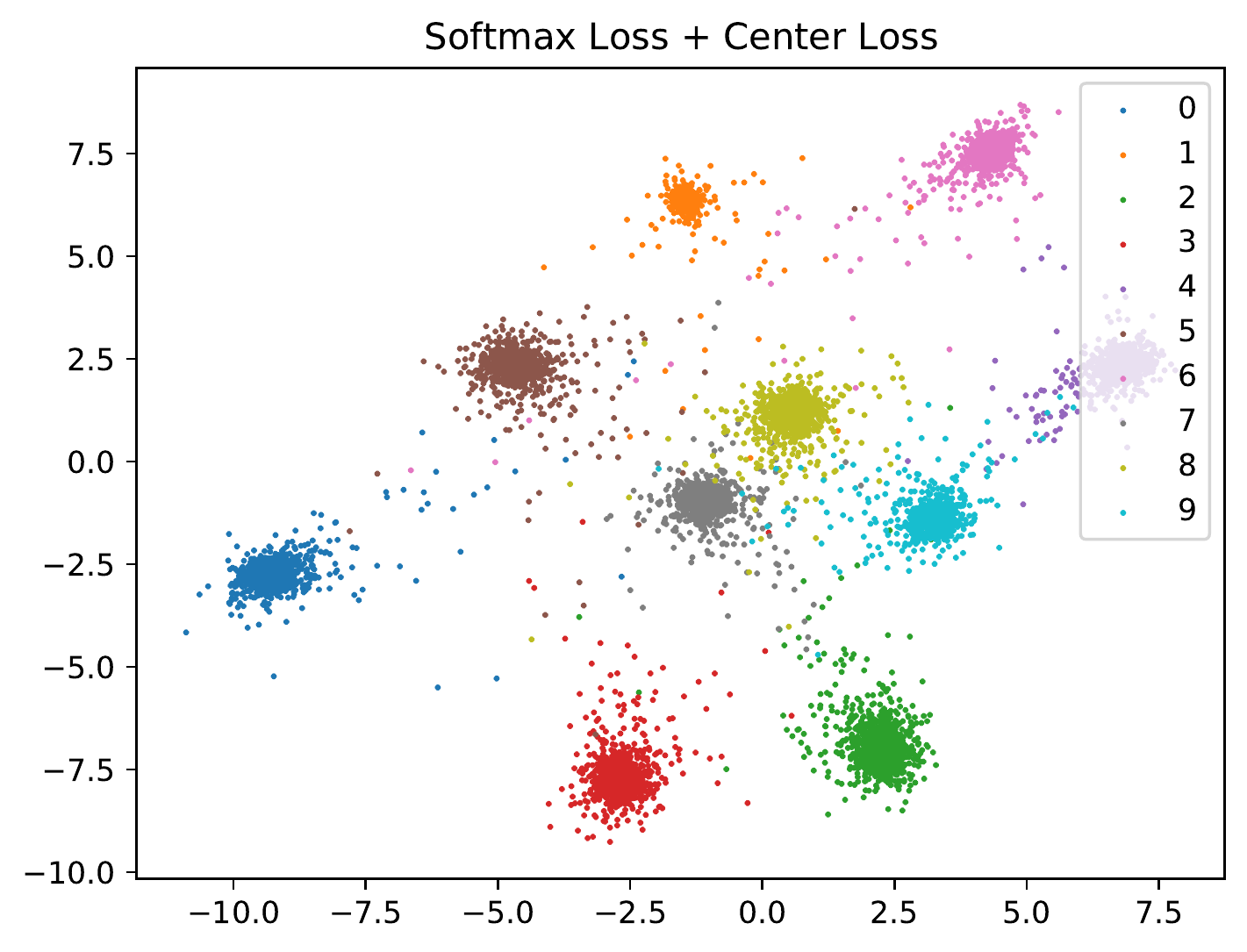} 
    \end{minipage}
    
\caption{Left: Visualization of features in 2D from a CNN model trained on MNIST dataset \cite{lecun1998mnist} with Softmax Loss. Right: Visualization of features in 2D from a CNN trained on MNIST dataset with joint supervision of Softmax Loss and Center Loss with \(\lambda\) = 1}
\label{fig:vis_mnist}
\end{figure}

\section{Discriminative Feature Learning with Center Loss}
\label{Discriminative_Feature_Learning_with_Center_Loss}

Typical deep neural network architecture comprises an input layer, followed by several hidden layers with non-linear activation functions and the output layer. The output layer generally has a softmax activation function for multi-class classification. This last fully connected layer acts as a linear classifier that separates the deeply learned features produced by the last hidden layer. The softmax loss forces the deep features of different classes to stay apart. The discriminative power of learned features is enhanced if the intra-class compactness and inter-class separability are maximized simultaneously. Though the features learned using the softmax loss are separable, they are not discriminative enough for open-set supervised problems and often exhibit high intra-class variance. This adversely affects the generalization capabilities of neural networks.

Several works \citep{wen2016discriminative, deng2017marginal, zhang2017range, liu2017sphereface, wang2018cosface, wang2018additive, chen2017noisy, wan2018rethinking, qi2018face} have proposed variants of softmax loss to enhance the discriminative power. The siamese network \cite{koch2015siamese} based approaches which use contrastive loss \cite{sun2015deep, hadsell2006dimensionality} and triplet loss \cite{schroff2015facenet}, learn the embeddings directly. These approaches face the problem of semi-hard sample mining and combinatorial explosion in the number of pairs or triplets, which significantly affect the effective model training \cite{deng2019arcface}. There are also angular margin penalty-based approaches that have shown significant improvements over softmax loss and have been explored in various directions, especially for large-scale face recognition \cite{liu2017sphereface, wang2018cosface, wang2018additive, liu2016large, deng2019arcface}.

\citet{wen2016discriminative} introduced the center loss for discriminative feature learning to solve deep face recognition. The joint supervision of softmax loss and center loss is used to obtain the inter-class dispersion and intra-class compactness by simultaneously learning the centers and minimizing the distances between the deep features and their corresponding class centers. The center loss has the same requirement as the softmax loss and needs no complex recombination of the training samples like contrastive loss and triplet loss which suffer from dramatic data expansion. The center loss is defined as follows:
\\

\begin{equation}
\\
\label{eq:center_loss}
L_c(x; \theta, c) = \frac{1}{2} \sum_{i=1}^{m} \|\mbox{f}_{L-1}(x_i; \theta) - c_{y_i}\|^2_2
\end{equation}

In Equation \ref{eq:center_loss}, the \(x_i\) denotes the \(i\)th sample, belonging to the \(y_i\)th class, \(c_{y_i} \in R^d \) denotes the \({y_i}\) th class center of deep features. The size of mini-batch and size of the feature dimension is \(m\) and \(d\), respectively. \(L\) is the total number of layers, and \(\mbox{f}_{L-1}\) is the feature vector of layer \(L-1\), which is just before the softmax classifier layer, and the \(\theta\) denotes the network parameters. 

The formulation effectively characterizes the intra-class variations.  In each iteration, the centers are computed by averaging the features of the corresponding classes. The deep features learned using the center loss are highly discriminative, clustered around the corresponding class centers, and linearly separable by the final fully connected layer, which acts as a linear classifier. Figure \ref{fig:vis_mnist} presents the visualizations of features obtained using softmax loss on the left and using joint supervision of softmax and center loss on the right. \citet{wen2016discriminative} provides detailed analysis and extensive experiments on center loss and its application in discriminative feature learning.

\end{document}